\documentclass{article}

\usepackage{arxiv}

\usepackage[utf8]{inputenc} 
\usepackage[T1]{fontenc}    
\usepackage{geometry}
\usepackage{hyperref}       
\usepackage{url}            
\usepackage{booktabs}       
\usepackage{amsfonts}       
\usepackage{nicefrac}       
\usepackage{microtype}      
\usepackage{lipsum}
\usepackage{xspace}
\usepackage{xcolor}
\usepackage{subcaption}
\usepackage{changepage}
\usepackage{fontawesome5}
\usepackage{enumitem}
\usepackage{csquotes}
\usepackage{adjustbox}
\usepackage{marginnote} 
\usepackage{xparse}
\usepackage{xifthen}
\usepackage{etoolbox}
\usepackage[table]{xcolor}
\usepackage{array}
\usepackage{listings}
\usepackage{float}
\usepackage{numprint}
\usepackage{fontawesome5}
\usepackage[most]{tcolorbox}
\usepackage{graphicx}
\graphicspath{ {./images/} }

\definecolor{KaggleBlue}{RGB}{35,191,255}

\newcommand{\tldr}[1]{%
    \begin{tcolorbox}[
        breakable,
        colback=gray!15,             
        colbacktitle=KaggleBlue!50,
        coltitle=black,              
        colframe=KaggleBlue!100!black,
        boxrule=0.45pt,               
        fonttitle=\ttfamily\normalsize\bfseries,
        fontupper=\small,
        title=TL;DR,
        top=2.2mm,                     
        bottom=2mm,                  
        enhanced,                    
        attach boxed title to top right={xshift=-4mm, yshift=-2.1mm},
        boxed title style={
            size=small,
            top=0.8mm, 
            bottom=0.6mm,
            left=6mm,
            right=6mm,
            colframe=KaggleBlue!100!black,
            colback=KaggleBlue!35
        }
    ]%
    #1
    \end{tcolorbox}
    \vspace{0.2cm}
}

\newcounter{kcitecounter}
\NewDocumentCommand{\kcite}{m m o}{%
  \leavevmode%
  \stepcounter{kcitecounter}%
  \def\kicon{}%
  \ifstrequal{#1}{notebook}{\def\kicon{\faFile*}}{}%
  \ifstrequal{#1}{dataset}{\def\kicon{\faDatabase}}{}%
  \ifstrequal{#1}{model}{\def\kicon{\faCogs}}{}%
  \texttt{[\href{#2}{\textcolor{KaggleBlue}{\kicon\hspace{3pt}\thekcitecounter}}]}%
}

\title{\textcolor{KaggleBlue}{Kaggle Chronicles:} 15 Years of Competitions, Community and Data Science Innovation}

\author{
  Kevin \xspace \textcolor{KaggleBlue}{\textsc{TheItCrow}} \xspace Bönisch \\
  \textsc{Goethe-University Frankfurt} \\
  \texttt{k.boenisch@outlook.com} \\
  \texttt{\href{https://www.kaggle.com/kevinbnisch}{kaggle.com/kevinbnisch}} \\
  \And
  Leandro \xspace \textsc{\textcolor{KaggleBlue}{BwandoWando}} \xspace Losaria \\ 
  \textsc{Pasig City, Philippines}\\
  \texttt{brando.losaria@gmail.com} \\
  \texttt{\href{https://www.kaggle.com/bwandowando}{kaggle.com/bwandowando}} \\
}

\begin{document}
\maketitle

\begin{center}
    \begin{tcolorbox}[
        colback=gray!10,
        colframe=KaggleBlue!100,
        arc=2mm, 
        boxsep=5pt, 
        boxrule=1pt,
        width=0.85\linewidth,
    ]
        \centering
        \small{\textit{Created as part of the \href{https://www.kaggle.com/competitions/meta-kaggle-hackathon}{Meta Kaggle Hackathon}}}\\
        \vspace{0.2cm} 
        \includegraphics[width=1.5cm]{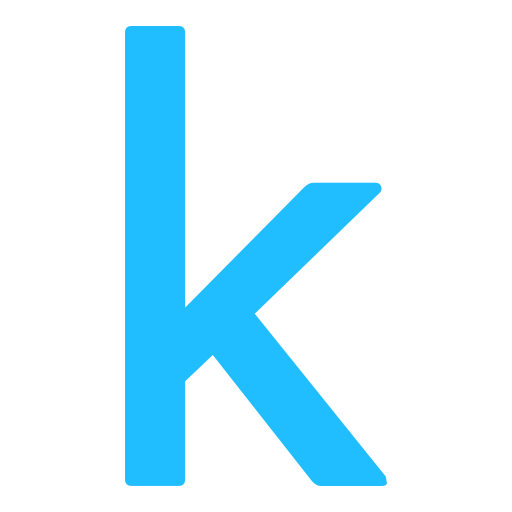}\\
        \vspace{0.2cm} 
        Team \textsc{\textcolor{KaggleBlue}{Bierbingka}}
    \end{tcolorbox}
\end{center}

\vspace{1cm}

\begin{abstract}
Since 2010, Kaggle has been a platform where data scientists from around the world come together to compete, collaborate, and push the boundaries of Data Science.
Over these 15 years, it has grown from a purely competition-focused site into a broader ecosystem with forums, notebooks, models, datasets, and more.
With the release of the Kaggle Meta Code and Kaggle Meta Datasets, we now have a unique opportunity to explore these competitions, technologies, and real-world applications of Machine Learning and AI.
And so in this study, we take a closer look at 15 years of data science on Kaggle--through metadata, shared code, community discussions, and the competitions themselves.
We explore Kaggle’s growth, its impact on the data science community, uncover hidden technological trends, analyse competition winners, how Kagglers approach problems in general, and more.
We do this by analyzing millions of kernels and discussion threads to perform both longitudinal trend analysis and standard exploratory data analysis.
Our findings show that Kaggle is a steadily growing platform with increasingly diverse use cases, and that Kagglers are quick to adapt to new trends and apply them to real-world challenges, while producing--on average--models with solid generalization capabilities.
We also offer a snapshot of the platform as a whole, highlighting its history and technological evolution.
Finally, this study is accompanied by a \href{https://www.youtube.com/watch?v=YVOV9bIUNrM}{video} and a \href{https://kaggle.com/competitions/meta-kaggle-hackathon/writeups/kaggle-chronicles-15-years-of-competitions-communi}{Kaggle write-up} for your convenience.

\end{abstract}


\newpage
\tableofcontents
\newpage

\section{Introduction}
\vspace{-0.5cm}
\tldr{
    We begin each section with a \texttt{TL;DR} (\textit{Too long; didn’t read}) box, which provides a rough summary of the insights presented in that section.
    These summaries are not intended to replace the full content, but rather to serve as a skim guide for identifying the most relevant sections.
}

\href{https://www.kaggle.com/code/nilaychauhan/your-search-ends-here-the-journey-of-kaggle}{Kaggle was founded in 2010 by Anthony Goldbloom and Ben Hamner} with the vision of creating a platform where data scientists from around the world could compete to develop the best possible solutions--measured by a specified metric--to a given problem within a limited timeframe, driven by the motivation of earning medals, recognition, and prize money.
Over time, Kaggle has evolved far beyond these competition-focused roots. With the introduction of dedicated discussion forums, integrated notebooks, public datasets, and most recently, model repositories, it has transformed into a comprehensive data science ecosystem.
This growth encouraged a continually expanding community, making Kaggle not only a place for competition but also a hub for learning, collaboration, and innovation in data science.
When Google acquired Kaggle in 2017, this development was further accelerated through greater integration with the Google ecosystem and infrastructure.

Meanwhile, the field of data science has undergone massive transformation.
Machine learning and AI exploded in popularity across industry, daily life, and academic research~\cite{liang-2024-mappingincreasingusellms, babina-2024-aigrowthinnovation}.
At the same time, the demand for computational resources has skyrocketed--particularly with the rise of (large) language models, which continue to grow in size and complexity~\cite{minaee-2025-largelanguagemodelssurvey}.
For example, recent studies show a dramatic increase in AI-generated content online, with a \numprint{57.3}\% rise on mainstream websites and an alarming \numprint{474}\% rise on misinformation sites~\cite{hanley-durumeric-2023-machine}, alongside growing concerns around plagiarism~\cite{bisi-etal-2023-rate, elali-rachid-2023-plagiarism, pudasaini-etal-2024-plagiarism}.
This trend is not limited to language technologies--AI adoption is also accelerating across healthcare, finance, entertainment, and transportation, powered by machine learning algorithms, vision models, and time series forecasting~\cite{chinimilli-2024-rise}.

Critically reflecting on this brings us to a central--yet often overlooked--necessity: one of the major challenges in managing a technological boom such as AI is the task of \textbf{accurately measuring and testing its capabilities}. 
That is, translating research and potential breakthroughs into rigorous, real-world application to answer questions such as:

\begin{center}
    \textit{How effective is AI for specific tasks? What are its boundaries--not just in theory, but in practice? And what real-world problems can it meaningfully and reliably solve?}
\end{center}

Blindly deploying state-of-the-art AI technologies in sensitive or high-stakes domains--such as financial forecasting, education, information retrieval, or medical diagnostics--without understanding their actual capabilities can lead to unintended and potentially harmful outcomes.

To mitigate this, researchers have turned to benchmark development~\cite{srivastava-2023-imitationgamequantifyingextrapolating, kim-2025-biggenbenchprincipledbenchmark}, controlled experiments, and systematic evaluations across domains~\cite{wang-2024-aicreativehumans, dave-2023-healthcare}.
Yet within this landscape, Kaggle stands out as a uniquely valuable case study: over the past fifteen years, it has gathered a global community that advances AI through open competitions, shared notebooks, public discussion, and collaborative experimentation.
With the release of Kaggle’s very own \textsc{meta-datasets}--Meta Kaggle and Meta Kaggle Code--we are now equipped to analyze Kaggle itself, providing a rare opportunity to reflect on how the platform and its community have contributed to the evolution of AI and Data Science as a whole.

And so in this study, we take a data-driven look at Kaggle’s development, examining how its users, forums, notebooks, and competitions have collectively helped shape the data science landscape--and how this insight can inform the future of empirically-driven AI development.
We follow a general research paper structure, but do not fully adhere to its conventions. Throughout this work, we frequently cite non-traditional sources--such as Kaggle notebooks (often enough created by us for this specific purpose, and cited like: \kcite{notebook}{https://kaggle.com}) or other forms of evidence--to support our claims.

\textbf{It is also important to note} that this work was conducted within the scope of the \href{https://www.kaggle.com/competitions/meta-kaggle-hackathon}{\texttt{Meta Kaggle Hackathon}}, under a limited time and resource frame.  
As such, we encourage readers to explore the cited resources directly to validate, extend, or build upon our findings and to acknowledge, that this is not a peer-reviewed scientific paper.

\newpage
\section{Kaggle User Growth}\label{sec:meta}
\tldr{
We've analyzed the growth of users in Kaggle over the years and have identified spikes in the daily counts in the registration of users. By identifying the spikes in new users registering, and not just the organic growth over time, we can gauge how influential Kaggle is in the datascience space and how effective is its democratization through associated events, competitions, and collaborations with other platforms. 
}

Kaggle was launched in April 2010 as a small platform hosting data science competitions, aiming to connect organizations with skilled individuals to address real-world problems and promote the democratization of data science. Over time, it evolved into a global community where users, regardless of background, can access datasets, share code, and learn from one another. Today, Kaggle plays a significant role in making data science more open and accessible by supporting skill development, career growth, and collaborative problem-solving. The platform also provides free access to high-performance tools such as GPUs, enabling users to run advanced models and experiments with ease.

By examining spikes in new user registrations, beyond the platform's steady organic growth, we can assess Kaggle’s influence within the data science community and evaluate how effectively it promotes democratization through key events, competitions, and partnerships with other platforms.

Some questions that we would like to answer with the analysis.

\begin{enumerate}
\item \textbf{When did key Kaggle membership milestones occur (e.g., 1M, 5M, 10M members)?}

\item \textbf{What specific dates show significant spikes in user registrations?}

\item \textbf{Which events, both within Kaggle and externally, contributed to these spikes?}

\item \textbf{How can the impact of these events be quantified by analyzing the number of users who registered during the associated periods?}

\end{enumerate}

\subsection{Early Years (2010-2015)}

\begin{figure}
    \centering
    \includegraphics[width=\linewidth]{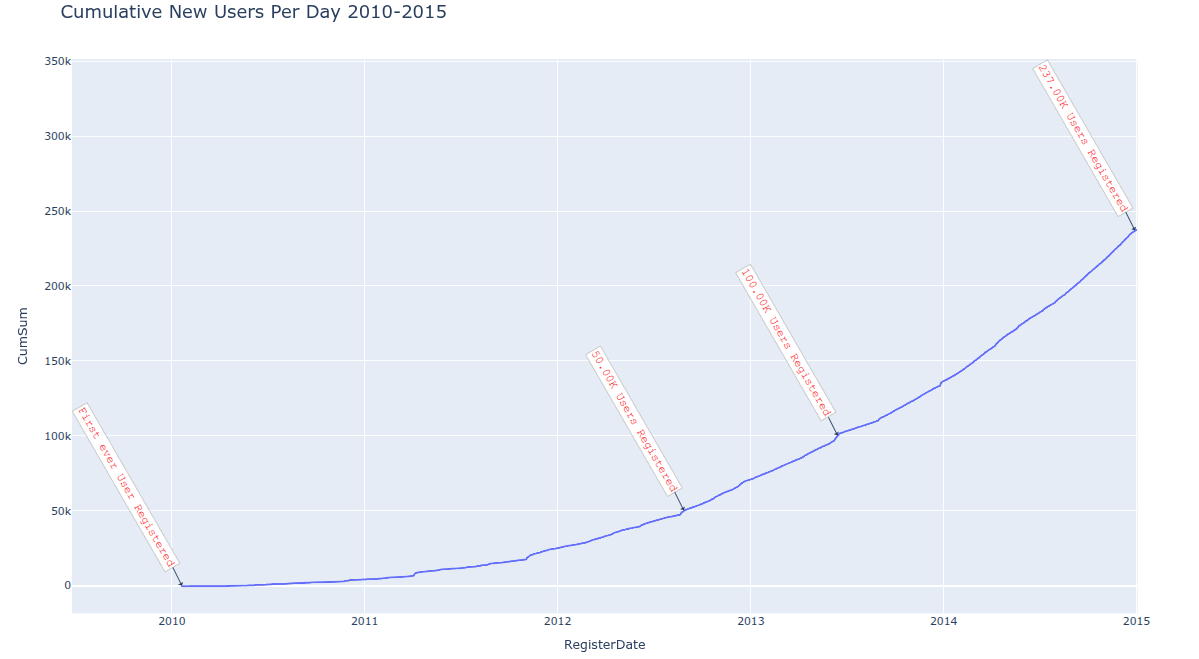}
    \caption{Kaggle Cumulative Growth 2010-2015 \kcite{notebook}{https://www.kaggle.com/code/bwandowando/kaggle-events-and-new-user-registration-counts}}
    \label{fig:growth-2010-2015}
\end{figure}

Kaggle began in April 2010 as a platform for data science competitions. It started small but quickly gained attention by allowing data science and machine learning practitioners to have a platform to discuss real-world problems, exchange ideas, and to learn from each other. Figure \ref{fig:growth-2010-2015}  shows that from 2010 to 2015, user registration remained relatively modest, with occasional spikes corresponding to competitions or media exposure. During this period, Kaggle was still building features, but already attracting attention from academia and early adopters. In general, this phase was defined by gradual organic growth driven by competition-based participation.

\subsection{Expansion and Pre-Pandemic Acceleration (2015-2020)}

\begin{figure}
    \centering
    \includegraphics[width=\linewidth]{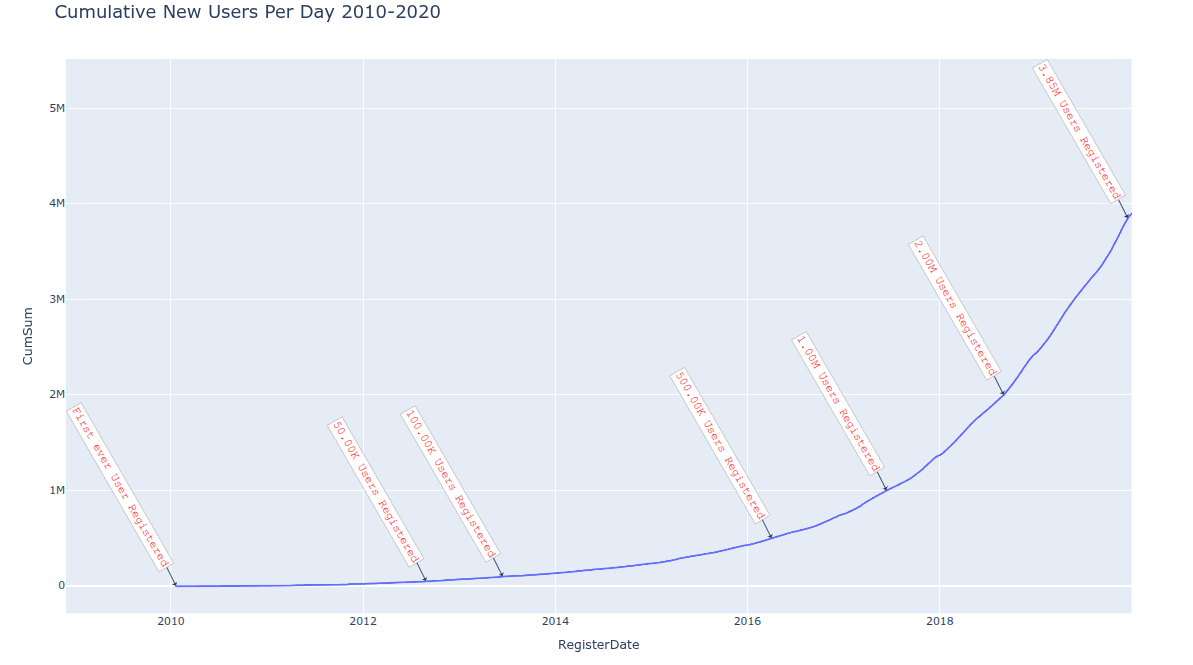}
    \caption{Kaggle Cumulative Growth 2010-2020 \kcite{notebook}{https://www.kaggle.com/code/bwandowando/kaggle-events-and-new-user-registration-counts}}
    \label{fig:growth-2010-2020}
\end{figure}

Figure \ref{fig:growth-2010-2020} shows that between 2015 and 2020, daily user registration began increasing more sharply during this period. A major milestone was the registration of the \textbf{one-millionth user in June 2017}, marking rapid adoption by learners, educators, and practitioners. Spikes in user registration during this phase indicate moments of high activity, likely tied to popular competitions, educational partnerships, and the launch of \href{https://www.kaggle.com/learn}{\textbf{Kaggle Learn}}. By late 2019, daily user growth had significantly increased,  as Kaggle slowly became mainstream, and offer free tools for data science education and collaboration. 

\subsection{COVID Pandemic era to Present (2020-present)}

Last March 2020, the \href{https://www.who.int/europe/emergencies/situations/covid-19}{\textbf{COVID-19 pandemic was announced by the World Health Organization}}. During the pandemic (highlighted in green on the graph), there was a surge in new user registrations. In a short span of just three years, from Mar 2020 to May 2023, the user count tripled from 4.2M to 13.3M. During this period, people around the world were likely stuck at home, looking for online learning opportunities and ways to upskill. Kaggle, being a platform for learning data science and machine learning, became a natural choice for these people. This COVID-era boom pushed Kaggle past 13M registered users, and even after the pandemic officially ended last May 2023, the growth trend remained strong.

Following the COVID-19 pandemic, Kaggle sustained its strong organic user growth. Starting in 2023, key events, such as the release of Gemma models in Kaggle, various newly introduced Large Language Model competitions, and hosting of Google's Gen AI intensive courses, coincided with spikes in daily user registration. With Kaggle's current growth, it is expected that by Q2 2026, Kaggle will reach \textbf{30 million registered users}, highlighting its continued relevance as a go-to platform for AI and data science innovation in the post-COVID digital landscape.

\begin{figure}
    \centering
    \includegraphics[width=\linewidth]{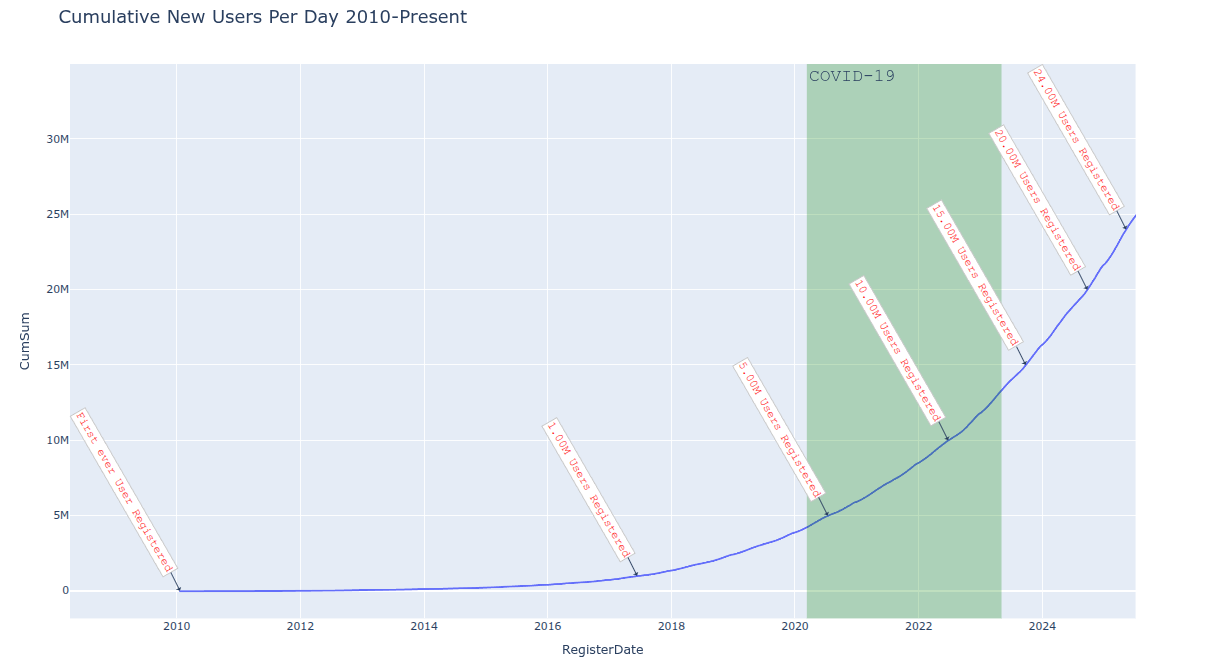}
    \caption{Kaggle Cumulative Growth 2010- Present \kcite{notebook}{https://www.kaggle.com/code/bwandowando/kaggle-events-and-new-user-registration-counts}}
    \label{fig:growth-2010-present}
\end{figure}

Figure \ref{fig:growth-2010-present} shows the COVID pandemic era and post-pandemic daily user registration counts of Kaggle.

\subsection{Anomaly Detection Using Sliding Window Z-test}

Anomaly detection using a sliding window Z-test is a statistically valid technique for identifying unusual spikes or drops in time series data, particularly in contexts like daily user registrations on platforms such as Kaggle. This method works by continuously analyzing recent historical data within a fixed-size window (we used 90 days), calculating the mean and standard deviation, and then assessing whether the next data point deviates significantly from the expected range using a Z-score. The formula can be seen below.

\begin{equation}
 \normalsize  Z\scriptsize score= \frac{\normalsize X\scriptsize  Registered Users On Observation Date - \normalsize\mu\scriptsize Registered Users Ninety Days Sliding Window}{\normalsize\sigma\scriptsize Registered Users Ninety Days Sliding Window}
\end{equation}

If a registration count falls far outside the typical variation observed in the sliding window, it is flagged as an anomaly, indicating potential external events, campaigns, or platform changes. This approach is especially relevant to Kaggle, where user activity can exhibit sharp, event-driven spikes in user registration due to competitions, events, and collaboration(s) with other platforms such as Coursera. 
\subsection{Identified Events and its effects on User Registration}

\begin{figure}
    \centering
    \includegraphics[width=\linewidth]{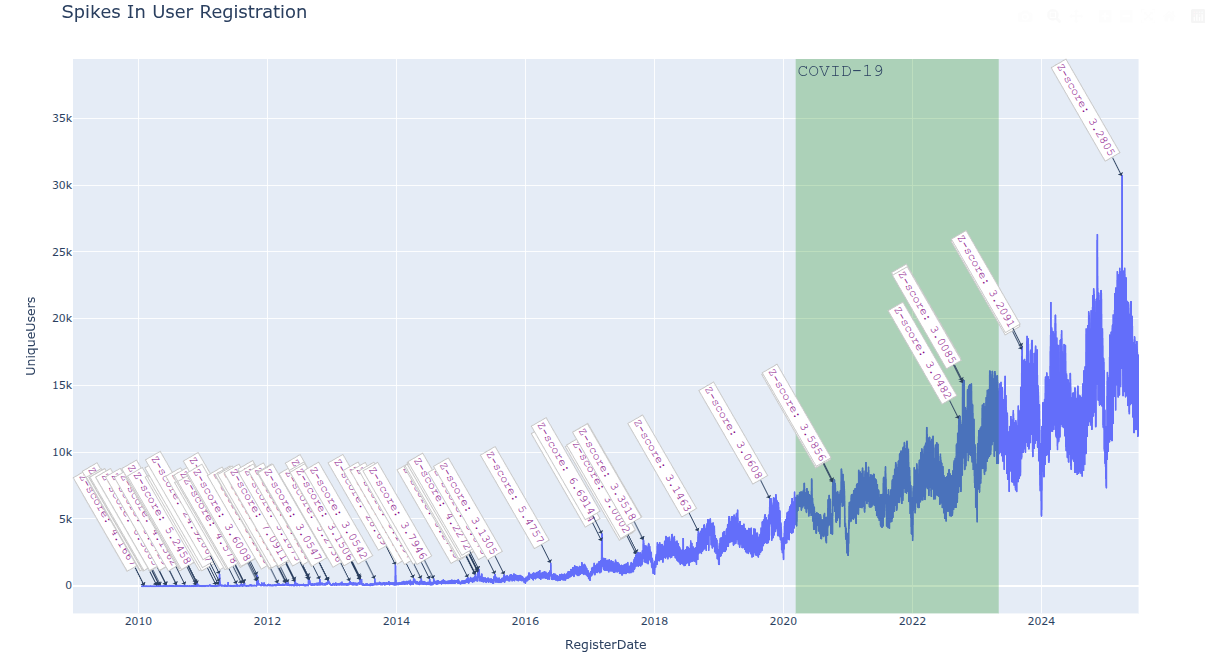}
    \caption{Spikes in Daily User Registration \kcite{notebook}{https://www.kaggle.com/code/bwandowando/kaggle-events-and-new-user-registration-counts}}
    \label{fig:spikes-in-registration}
\end{figure}

Although Kaggle's overall growth has been largely organic, Figure \ref{fig:spikes-in-registration} reveals notable spikes in daily user registrations. We've investigated and extensively looked into the MetaKaggle dataset, with the aid of Google search and ChatGPT, and validated the events that we've associated these spikes with. The dataset can be found \href{https://kaggle.com/datasets/e563a2df7847f2b5712c96e06e733937f1495a174efbed2d7e167076c467d8ee}{here}. We have associated events that coincided with these dates if an event was held within +/- 2 days of these spikes in user registration. 

Having calculated the Z-scores of the daily registration counts (see Table \ref{tab:z-scores}), we can see multiple dates that have a Z-score of more than 2.5, suggesting that the effect of these events within and outside Kaggle was not due to random chance. 

\renewcommand{\arraystretch}{1.3} 
\rowcolors{2}{gray!10}{white}

\begin{table}
\caption{\vspace{0.2cm}Kaggle events with the highest Z-scores}
\label{tab:z-scores}
\centering
\begin{adjustbox}{width=\textwidth}
\begin{tabular}{@{}r l l r r r@{}}
\toprule
\textbf{Rank} & \textbf{Date} & \textbf{Event} & \textbf{Registered Users} & \textbf{90-day Rolling Window Daily Average} & \textbf{Z-score} \\
\midrule
1  & 2013-12-26 & \href{https://www.kaggle.com/competitions/packing-santas-sleigh}{Packing Santa's Sleigh} & 1578  & 192.177778  & 28.6900 \\
2  & 2011-04-05 & \href{https://www.kaggle.com/competitions/hhp}{Heritage Health Prize} & 889   & 30.133333   & 24.9206 \\
3  & 2011-11-04 & \href{https://www.finsmes.com/2011/11/kaggle-raises-11m-series-funding.html}{Kaggle Raises \$11M in Series A Funding} & 803 & 51.977778 & 15.5391 \\
4  & 2012-08-22 & \href{https://www.kaggle.com/competitions/predict-closed-questions-on-stack-overflow}{Predict Closed Questions on Stack Overflow} & 670 & 103.066667 & 8.3101 \\
5  & 2017-03-08 & \href{https://www.kaggle.com/discussions/general/29781}{Kaggle is joining Google Cloud!} & 3353 & 1215.111111 & 6.5344 \\
6  & 2014-04-08 & \href{https://publichealth.jhu.edu/2014/coursera-specialization}{Johns Hopkins Specialization in Data Science Through Coursera} & 580 & 245.833333 & 4.9852 \\
7  & 2015-04-08 & \href{https://www.kaggle.com/c/march-machine-learning-mania-2015}{March Machine Learning Mania 2015} & 984 & 469.977778 & 3.3669 \\
8  & 2022-10-11 & \href{https://www.kaggle.com/competitions/google-universal-image-embedding}{Google Universal Image Embedding} & 15361 & 8473.200000  & 3.3053 \\
9  & 2025-03-31 & \href{https://www.kaggle.com/learn-guide/5-day-genai}{5-Day Gen AI Intensive Course with Google (Mar 2025)} & 30701 & 17176.144444 & 3.2805 \\
10  & 2010-11-23 & \href{https://www.kaggle.com/competitions/RTA}{RTA Freeway Travel Time Prediction} & 43 & 12.888889 & 3.1082 \\
11 & 2024-02-22 & \href{https://www.kaggle.com/code/nilaychauhan/fine-tune-gemma-models-in-keras-using-lora}{Fine-tune Gemma models in Keras using LoRA} & 21223 & 12483.444444 & 2.8890 \\
12 & 2024-11-11 & \href{https://rsvp.withgoogle.com/events/google-generative-ai-intensive}{5-Day Gen AI Intensive Course with Google (Nov 2024)} & 26313 & 15929.644444 & 2.7808 \\
13 & 2020-11-23 & \href{https://www.kaggle.com/competitions/jane-street-market-prediction}{Jane Street Market Prediction} & 8741 & 5793.444444 & 2.4862 \\
14 & 2023-10-10 & \href{https://www.kaggle.com/competitions/ai-village-capture-the-flag-defcon31}{AI Village Capture the Flag @ DEFCON31} & 18685 & 12046.111111 & 2.3507 \\
15 & 2018-04-10 & \href{https://www.kaggle.com/competitions/data-science-bowl-2018}{2018 Data Science Bowl} & 3875 & 2760.411111 & 2.0961 \\
16 & 2021-07-31 & \href{https://www.kaggle.com/thirty-days-of-ml}{30 Days of ML} & 8915 & 6705.400000 & 2.0519 \\
17 & 2019-11-19 & \href{https://www.kaggle.com/competitions/link-prediction-data-challenge-2019}{Link Prediction Data Challenge} 2019 & 6823 & 4620.333333 & 2.0212 \\
18 & 2016-11-14 & \href{https://www.kaggle.com/c/the-nature-conservancy-fisheries-monitoring/}{The Nature Conservancy Fisheries Monitoring} & 1574 & 1067.311111 & 2.0017 \\
19 & 2018-11-12 & \href{https://www.kaggle.com/competitions/inclusive-images-challenge}{Inclusive Images Challenge} & 5027 & 3453.133333 & 1.9391 \\
20 & 2016-05-24 & \href{https://www.kaggle.com/competitions/can-we-predict-voting-outcomes}{Can we predict voting outcomes?} & 1197 & 849.455556 & 1.8972 \\
\bottomrule
\end{tabular}
\end{adjustbox}
\end{table}

\begin{figure}
    \centering
    \includegraphics[width=\linewidth]{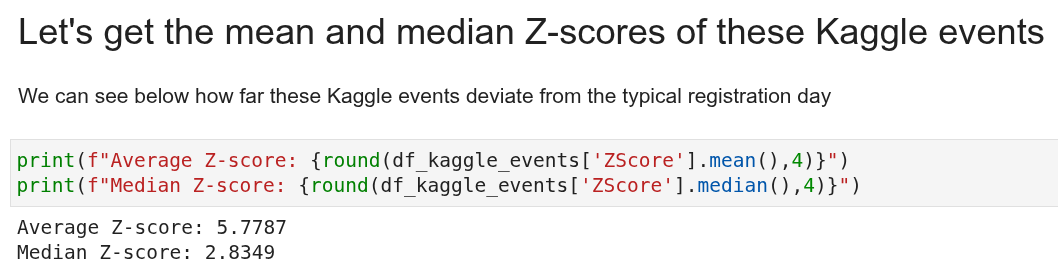}
    \caption{Mean and Median Z-scores\kcite{notebook}{https://www.kaggle.com/code/bwandowando/kaggle-events-and-new-user-registration-counts}}
    \label{fig:events-and-zscores}
\end{figure}

A higher Z-score indicates a day with significantly more registrations than typical. In Figure \ref{fig:events-and-zscores}, the mean \textbf{Z-score: 5.7787} reflects the overall intensity of these registration surges, suggesting that, on average, these events deviate substantially from normal activity. The median \textbf{Z-score: 2.8349}, being slightly lower, implies that while some events are extreme outliers, the majority are more moderately above the norm. 

\begin{figure}
    \centering
    \includegraphics[width=\linewidth]{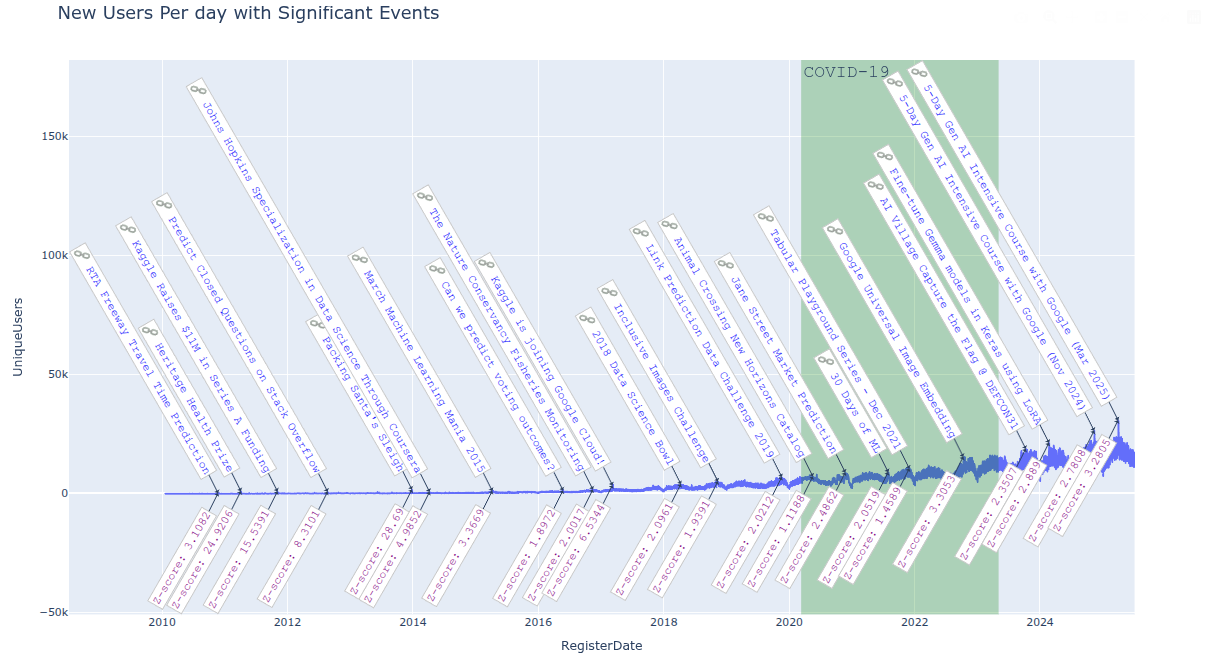}
    \caption{Registration Spikes, with Events and Z-scores \kcite{notebook}{https://www.kaggle.com/code/bwandowando/kaggle-events-and-new-user-registration-counts}}
    \label{fig:events-and-zscores-over-time}
\end{figure}

 In Figure \ref{fig:events-and-zscores-over-time} we've super-imposed the Z-values againt the spikes in registration over time would yield the next graph, and with this graph, we can now clearly see the effect of these events on registrations.

\subsection{Conclusion}

\begin{figure}
    \centering
    \includegraphics[width=\linewidth]{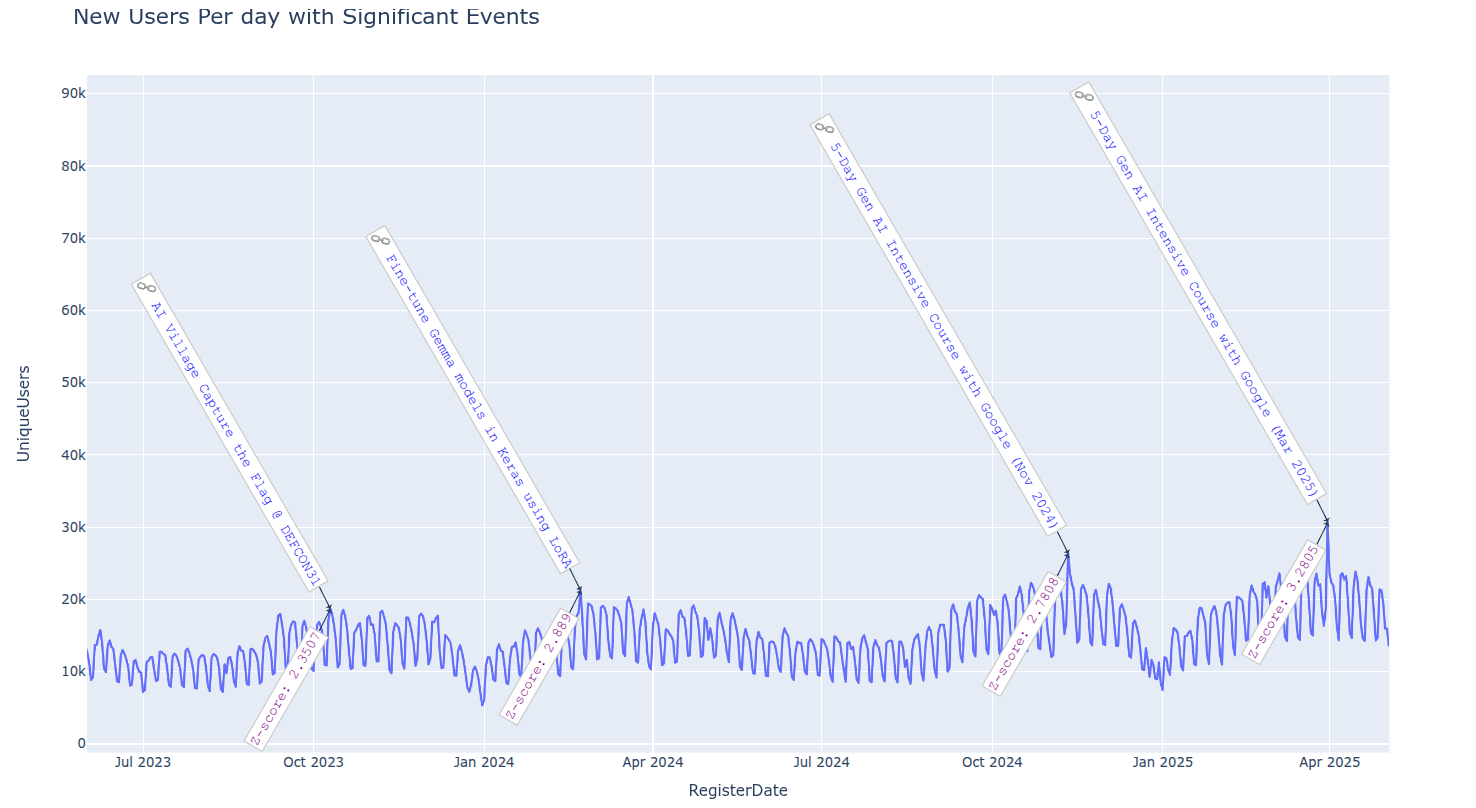}
    \caption{5-Day Gen AI Intensive Course with Google (Mar 2025) \kcite{notebook}{https://www.kaggle.com/code/bwandowando/kaggle-events-and-new-user-registration-counts}}
    \label{fig:5-day-gen-ai}
\end{figure}

Using simple statistical checks, whenever there is an official Kaggle event or collaboration with another platform, there were significant increases in new user sign-ups on Kaggle. We can also quantify the effect and reach of Kaggle during these events through counts in new users! These events are often linked to competitions, online courses, or collaborations with schools and companies, \textbf{with the \href{https://www.kaggle.com/learn-guide/5-day-genai}{5-Day Gen AI Intensive Course with Google} attracting the most number of new users in single day with 30K users!} Figure\ref{fig:5-day-gen-ai}.

Since Kaggle’s mission includes spreading data science knowledge to a wider audience, it's not surprising that such events draw more people to the platform. This pattern shows that the public responds well to opportunities to learn or take part in challenges. The spikes in registrations are a clear sign that Kaggle events help attract and involve more users, helping to make data science more accessible to everyone.

We also saw a clear rise in new Kaggle users during the COVID-19 period, when many people were staying at home. With more free time and a need to stay productive, many turned to learning new skills. Kaggle became a go-to place for this. Kaggle offered not just the tools and data, but also the community and the platform for people to grow their data science knowledge and connect with other practitioners around the world.

These findings highlight how far Kaggle’s influence has grown within the global data science community. These insights underscore Kaggle’s significant role and evolving influence in the data science domain. In its early years, Kaggle began primarily as a competition platform, over time, it has grown into a comprehensive platform that supports not only competitions but also learning, collaboration, and career development. Its ability to engage users, whether through partnerships, educational programs, or timely challenges, demonstrates its effectiveness in making data science more approachable and inclusive. Its continued growth reflects not just its technical offerings, but also its success in democratizing access to data science and empowering individuals to advance in this rapidly evolving field.
\newpage
\section{Investigating Meta Kaggle Code}
\vspace{-0.5cm}
\tldr{
    Kaggle's kernels are predominantly written in Python, with occasional usage of R.  
    We investigate package imports and method calls, and find a diverse range of packages and technologies being used, both in out-of-competition and competition kernels.  
    The entropy and overall diversity of these technologies have been increasing year by year.  
    Specifically in competitions, we observe a rapid adaptation to new technologies, some of which even dominate the winners' write-ups.  
}

The \href{https://www.kaggle.com/datasets/kaggle/meta-kaggle-code?_gl=1*1uh8axr*_ga*MTYxMjg4MTI5LjE3MzI5MTU4NTE.*_ga_T7QHS60L4Q*czE3NDgwMzU5MTUkbzE3NyRnMSR0MTc0ODAzNjI1NyRqMCRsMCRoMA..}{\textsc{Meta Kaggle Code}} dataset provides an extensive archive of publicly shared notebooks (\textit{kernels}) created by users on the Kaggle platform. 
It contains approximately \numprint{5900000} notebooks, each accompanied by metadata such as author information, timestamps, competition status, and the full source code.  
This dataset covers a time span from 2015 to the present, amounting to roughly \numprint{300} gigabytes of data. Its size and diversity make it a valuable resource for understanding trends in data science practice, particularly in the context of real-world, user-generated content.  
Unlike many benchmark datasets that focus solely on outputs or results, Meta Kaggle Code enables the analysis of actual codebases at scale - something that is often neglected.
This opens the door to exploring how data scientists structure their workflows, which tools and methods they prefer, and how coding behavior evolves over time.  

So in this section, we take it further than a regular metadata analysis and examine the code itself. We do this by parsing import statements, method calls, and library usage patterns; we identify technological trends and behavioral patterns in the Kaggle community.
We distinguish between code written for competitions and that created for general experimentation or sharing, enabling us to compare these two modes of engagement.  
This approach allows us to address several key research questions:

\begin{enumerate}
    \item \textbf{Which packages and libraries are most frequently used on Kaggle?} Are there dominant tools, and how do their usage patterns evolve over time?
    \item \textbf{Which technologies are most prominent in competitions?} Do certain frameworks or methods correlate with higher performance or winning submissions?
    \item \textbf{How diverse is the Kaggle ecosystem in terms of programming technologies?} Can we quantify technological diversity or detect shifts toward particular toolchains?
    \item \textbf{How fast do kagglers adapt?} Are they working with cutting-edge technology? If so, how effective? And consequently: which tools stand the test of time and stay?
    \item \textbf{What technologies do winning Kagglers rely on?} Do winners converge around certain best practices, or do they innovate with lesser-known tools?
\end{enumerate}

Each of the conducted data analyses has been made publicly available by us in the form of notebooks and source code, for your convenience and to ensure proper citation.
All of the facts and insights presented in this work were generated based on data available as of June 2025.  
Accordingly, statements referring to the current state of the platform, such as \enquote{as of now} should be understood in the context of that time frame.

\subsection{R or Python?}

\tldr{
    Since 2016, the use of the R programming language on Kaggle has been steadily declining, while Python has seen continuous growth.  
    As of now, approximately 95\% of all Kaggle kernels are written in Python \kcite{notebook}{https://www.kaggle.com/code/kevinbnisch/python-or-r-what-do-kagglers-use}.
}

\begin{figure}
    \centering
    \includegraphics[width=\linewidth]{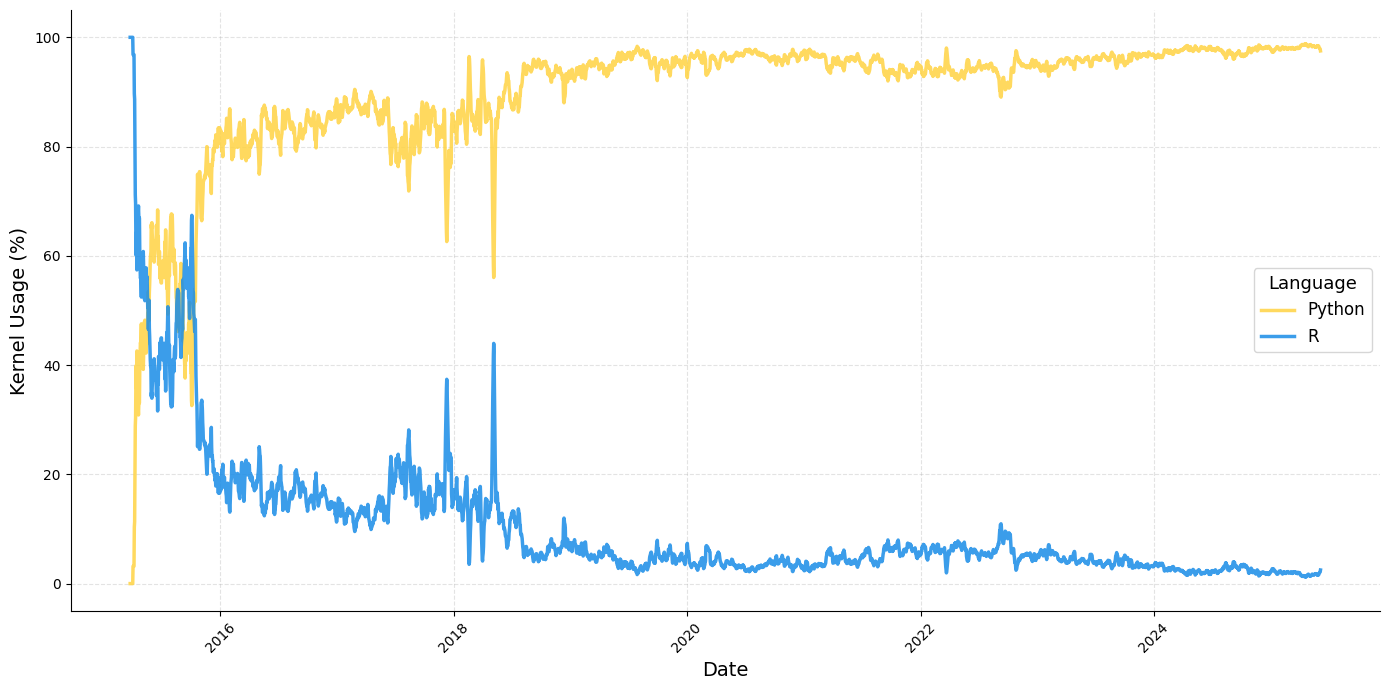}
    \caption{The development of R versus Python kernels on Kaggle reveals a clear and consistent trend toward Python since 2016.}
    \label{fig:r-vs-python}
\end{figure}

Out of the total of \numprint{1487483} kernels, approximately 5\% are written in R, while the remaining 95\% are written in Python.  
The earliest kernel in the dataset dates back to 2015-03-25, and--as shown by Figure \ref{fig:r-vs-python}--Python usage has steadily increased in popularity since then.  

The most upvoted R kernel belongs to a user named \textit{chebotinaa}. 
Their R \href{https://www.kaggle.com/code/chebotinaa/bellabeat-case-study-with-r}{kernel}, paradoxically created in 2021, has received a total of 728 upvotes. 
The user with the highest number of R kernels is \textit{willianoliveiragibin}, with an impressive total of \numprint{332} R kernels.  

In comparison, \textit{alexisbcook} authored the Python \href{https://www.kaggle.com/code/alexisbcook/titanic-tutorial}{kernel} with the highest number of upvotes: \numprint{17847}.  
Unsurprisingly, this kernel is part of the well-known Titanic Tutorial series, serving as an introductory notebook for newcomers to Kaggle.

\subsection{What Packages \& Methods do Kagglers use in General?}

\tldr{
    The most frequently imported packages on Kaggle are \texttt{pandas}, \texttt{numpy}, and \texttt{matplotlib}, particularly the \texttt{matplotlib.pyplot} module.  
    Among R kernels, the most commonly imported package is \texttt{ggplot2}, with over \numprint{100000} imports.  
    These prominent packages often appear together, indicating a strong co-dependency and forming a stable foundation for many Kaggle workflows.  
    When combined with the observed method usage distribution, we can infer how data science is practiced on Kaggle: a pragmatic, modeling-driven approach, yet with a focus on visualization and frequent basic data cleaning.
}

We analyzed all kernel versions available in the \textsc{Meta Code} dataset and extracted package imports and method calls from \numprint{5357315} kernels by using regex commands \kcite{notebook}{https://www.kaggle.com/code/kevinbnisch/most-used-packages-methods-on-kaggle}.
Since this computation is resource-intensive, we make the resulting data publicly available as a dataset for your convenience~\kcite{dataset}{https://www.kaggle.com/datasets/kevinbnisch/kaggles-most-used-packages-and-method-calls}.

\begin{figure}[htbp]
  \centering
  \begin{minipage}[b]{0.49\textwidth}
    \centering
    \includegraphics[width=\textwidth]{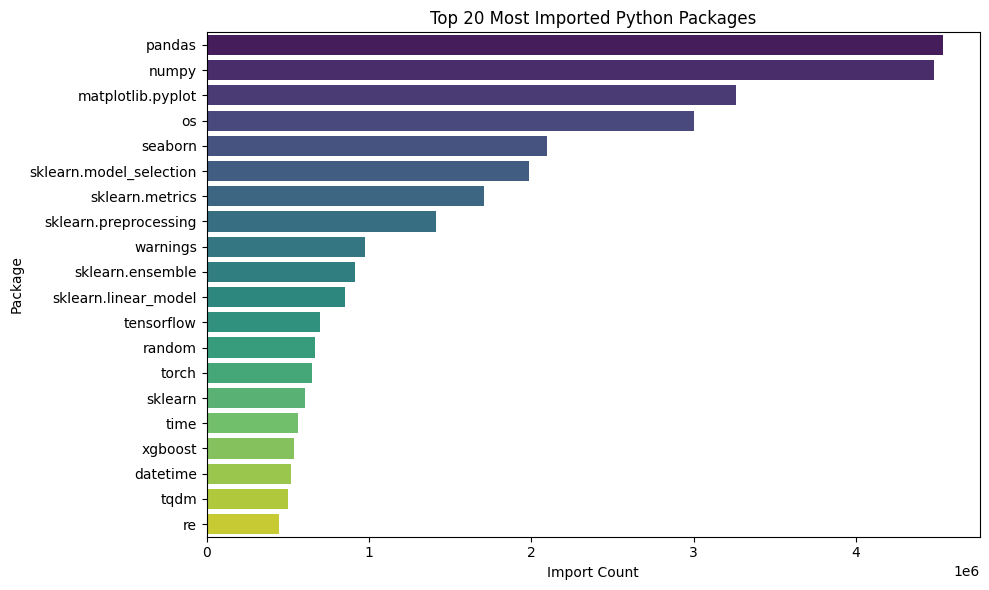}
  \end{minipage}
  \hfill
  \begin{minipage}[b]{0.49\textwidth}
    \centering
    \includegraphics[width=\textwidth]{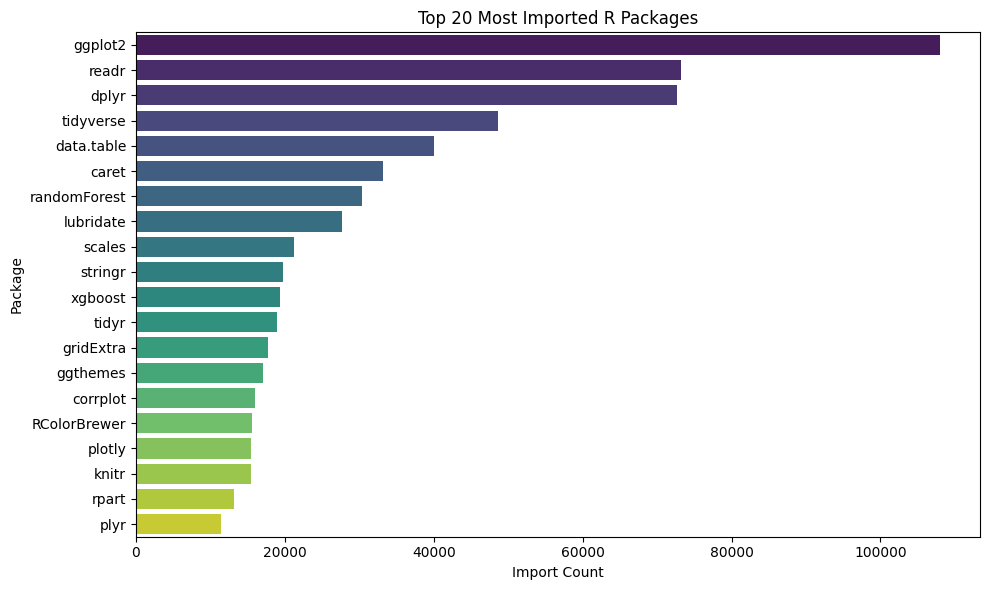}
  \end{minipage}
  \caption{The top 20 overall most imported packages, left for Python and right for R \kcite{notebook}{https://www.kaggle.com/code/kevinbnisch/most-used-packages-methods-on-kaggle}.}
  \label{fig:side_by_side}
\end{figure}

When looking at Figure \ref{fig:side_by_side}, the total number of Python package imports is nearly two orders of magnitude higher than that of R packages (as a consequence of Python's dominance on the platform).  
It is no surprise that \texttt{pandas} and \texttt{numpy} are the most frequently imported packages, as they are commonly included at the start of nearly every Python kernel.  
Perhaps surprisingly, the third most imported package is \texttt{matplotlib.pyplot}, rather than the base namespace \texttt{matplotlib}.
This is likely because most of Matplotlib’s functionality being concentrated within the \texttt{pyplot} submodule, which is typically imported using the alias \texttt{plt}. This submodule provides commonly used functions such as the widely used \texttt{plot()} method.
On the R side, the most prominent packages are \texttt{ggplot2}, \texttt{readr}, and \texttt{dplyr}.

\begin{figure}
    \centering
    \includegraphics[width=\linewidth]{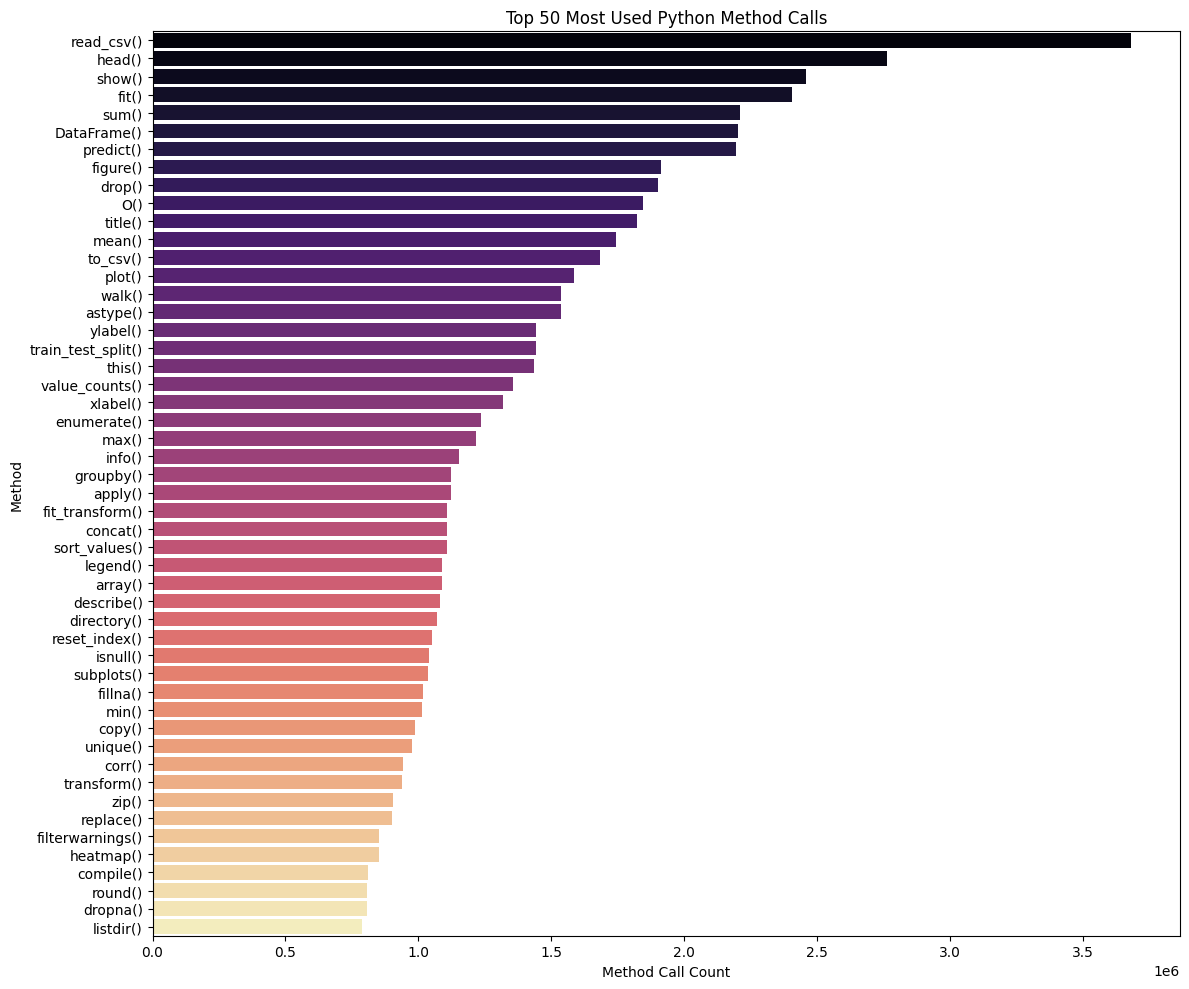}
    \caption{The top 50 most called methods in all Python Kernels on Kaggle \kcite{notebook}{https://www.kaggle.com/code/kevinbnisch/most-used-packages-methods-on-kaggle}.}
    \label{fig:top-used-methods}
\end{figure}

For Python kernels, we also extracted and analyzed method calls.  
Figure~\ref{fig:top-used-methods} shows the 50 most frequently used methods. As expected, the list is headed by \texttt{read\_csv()}, the typical entry point for most data science workflows. This method is provided by \texttt{pandas}, as well as by other libraries such as \texttt{polars}.  
It is followed by \texttt{head()}, commonly used to preview a dataframe, and \texttt{show()}, typically used to render plots or outputs.  
Interestingly, model-related method calls such as \texttt{fit()} and \texttt{predict()} appear more frequently than classic preprocessing methods like \texttt{drop()}, \texttt{apply()}, or \texttt{fillna()}, suggesting a strong focus on modeling possibly even at the early stages of many kernels—maybe even at the risk of working with less clean data?

This modeling-centric trend is further supported by the prominent appearance of \texttt{train\_test\_split()} and \texttt{fit\_transform()}, both core elements of machine learning workflows. It suggests that a significant portion of Kaggle users quickly transition from data loading to model training, possibly due to the competition-driven, results-oriented nature of the platform.

At the same time, plotting functions such as \texttt{plot()}, \texttt{figure()}, \texttt{title()}, \texttt{xlabel()}, \texttt{ylabel()}, and \texttt{legend()} are also highly represented, indicating that visualization plays a crucial role in EDA and communication of results. This underlines our assumption that the typical Kaggle kernel is not just about modeling, but also about presenting findings in a visually interpretable way.

The inclusion of \texttt{dropna()}, \texttt{fillna()}, and \texttt{isnull()} highlights the importance of dealing with missing values, though their relatively lower rank compared to modeling methods might suggest that data cleaning is sometimes treated as secondary—or at least performed using fewer unique calls.
This would align with our previous findings.

Lastly, the presence of \texttt{filterwarnings()} implies a degree of attention to notebook readability and presentation quality. Users may suppress warnings to beautify output, especially when preparing kernels for public sharing or competition submission.

Overall, the method usage distribution offers an interesting look into how data science is practiced in Python on Kaggle: \textbf{a pragmatic, modeling-driven approach, supported by visualization and basic data cleaning.}

\subsubsection{Over Time}

\begin{figure}[htpb]
  \centering
  \begin{minipage}[b]{\textwidth}
    \centering
    \includegraphics[width=\textwidth]{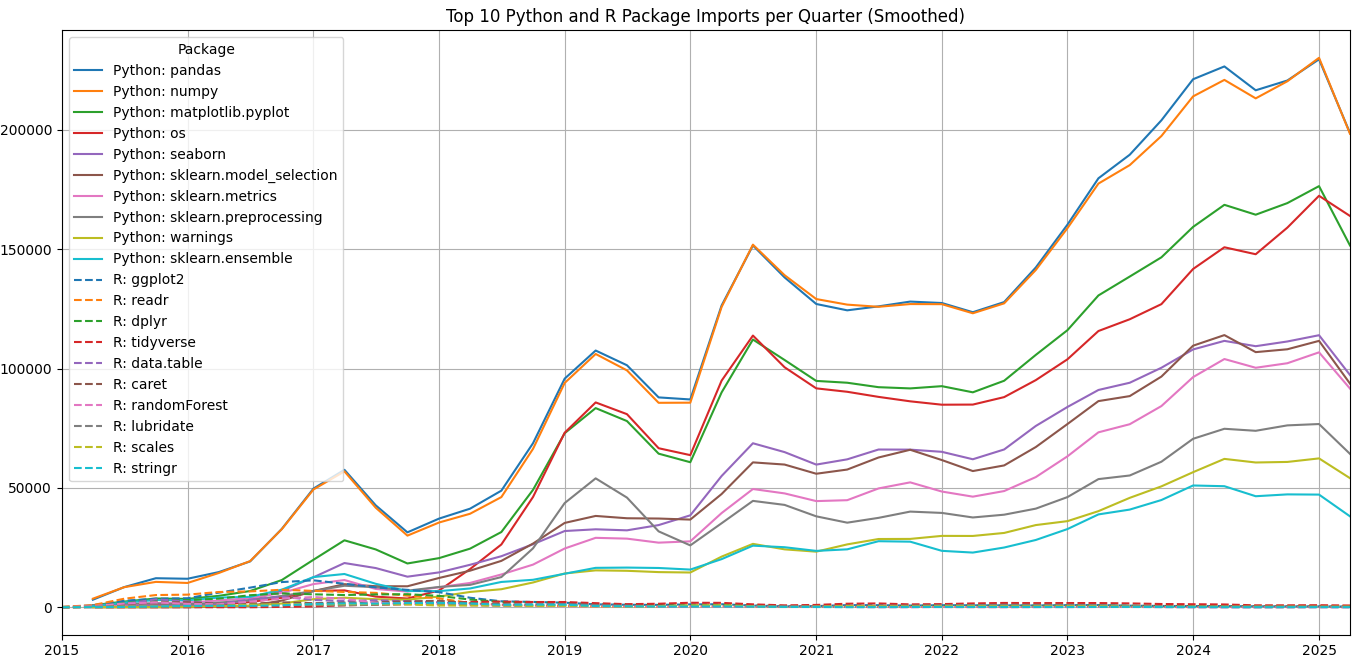}
  \end{minipage}
  \vspace{0em}

  \begin{minipage}[b]{0.55\textwidth}
    \centering
    \includegraphics[width=\textwidth]{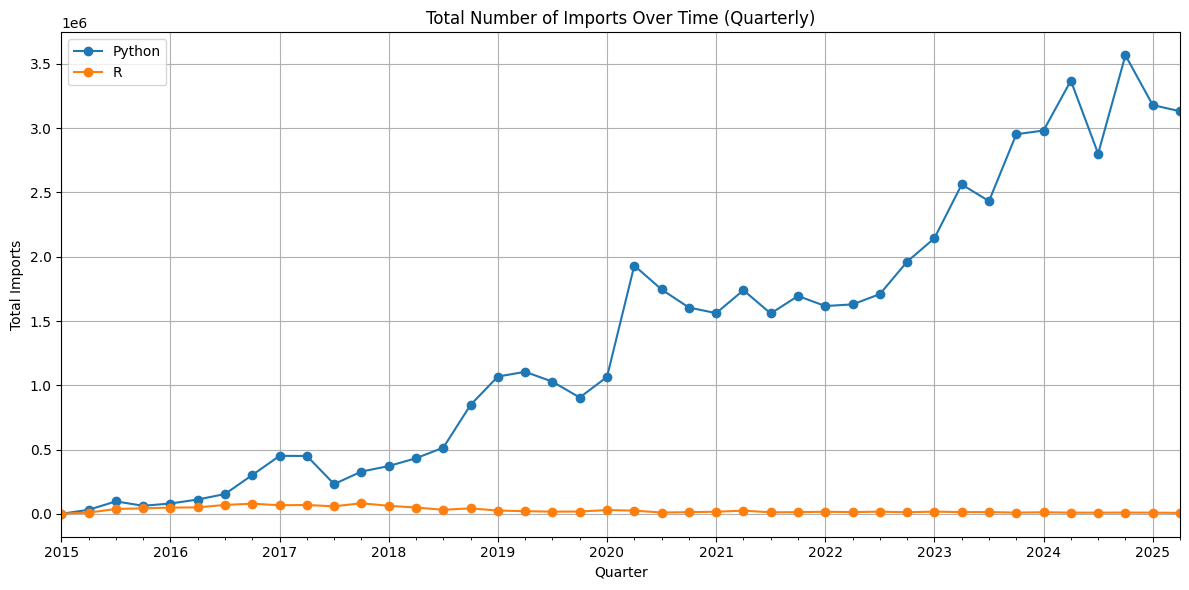}
  \end{minipage}
  \hfill
  \begin{minipage}[b]{0.44\textwidth}
    \centering
    \includegraphics[width=\textwidth]{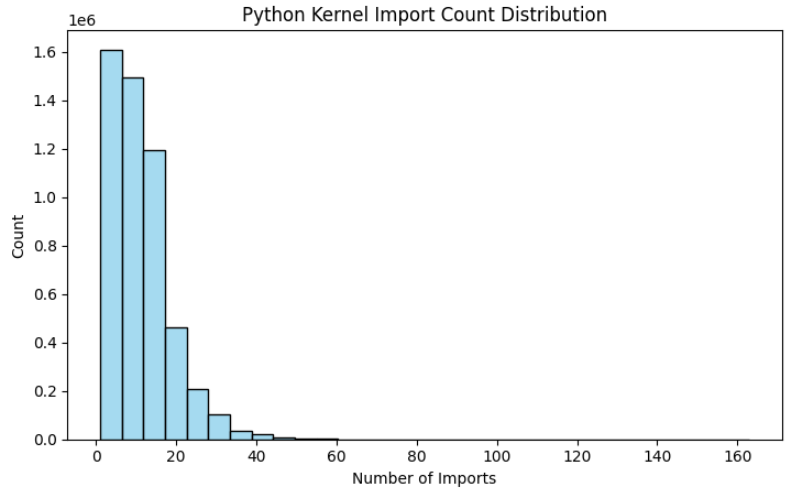}
  \end{minipage}

  \caption{The top 10 most imported packages overall, for both Python and R. Below that, the total number of package imports over time is shown for both languages (left), along with the distribution of the number of imported packages per Python kernel (right) \kcite{notebook}{https://www.kaggle.com/code/kevinbnisch/most-used-packages-methods-on-kaggle}.
  }
  \label{fig:packages-over-time}
\end{figure}

We now turn to an analysis of package usage over time, as presented in Figure~\ref{fig:packages-over-time}.  
At first glance, the top chart reveals two key observations:

\begin{enumerate}[leftmargin=0pt, labelindent=0pt, labelsep=1em, itemindent=0pt]
    \item The most frequently used packages—particularly in Python—show strong similar usage patterns, suggesting an interdependency.  
    This pattern likely reflects a default toolbox commonly utilized by Kaggle users and data scientists for a wide variety of tasks.
    
    \item Secondly, the total number of Python package imports increases steadily over time, while R imports gradually decline.  
    This trend is also visible in the bottom-left plot (showcasing the total imports over time) and can be attributed to the overall rise in Kaggle’s popularity and user engagement (see Section~\ref{sec:meta}), which has led to an increasing number of submitted kernels.
\end{enumerate}

The bottom-right plot shows the distribution of the number of imports per kernel. Most kernels include approximately 2 to 5 imports, with the frequency gradually decreasing in a roughly Gaussian fashion toward higher counts, reaching up to 40 or more imports.
This distribution suggests that Kaggle users typically rely on a minimal set of tools when writing kernels, often importing only the essential libraries required for data loading, manipulation, and visualization.  
The tapering tail toward 40 or more imports indicates the presence of more complex workflows, although these remain relatively rare.  
Overall, this indicates a tendency toward concise and narrowly scoped kernels, which supports simplicity, ease of understanding, and reproducibility--especially when kernels are forked or reused by other users.

\subsubsection{By Category}

\begin{figure}
  \centering

  \begin{subfigure}[b]{0.49\textwidth}
    \centering
    \includegraphics[width=\textwidth]{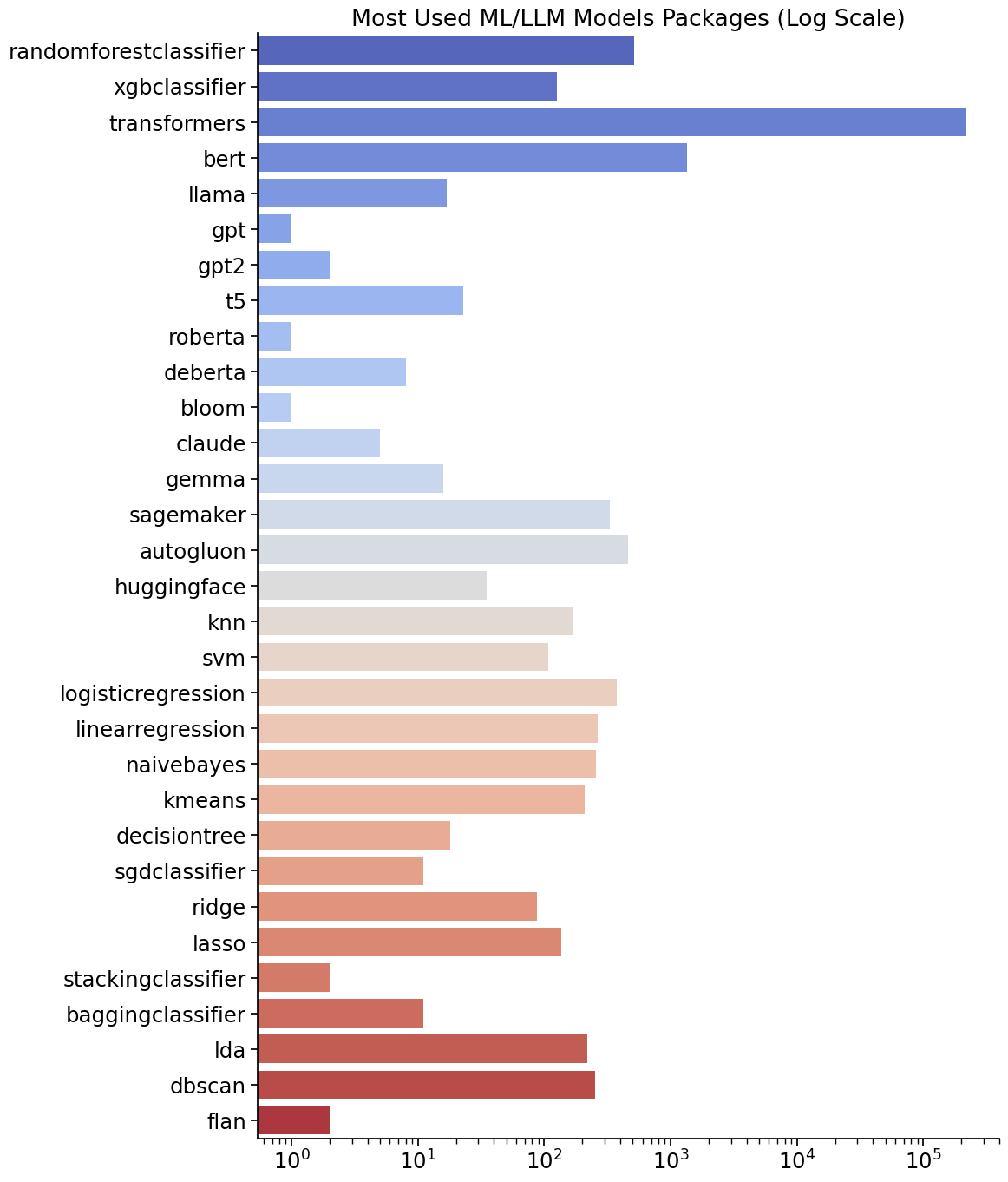}
  \end{subfigure}
  \hfill
  \begin{subfigure}[b]{0.49\textwidth}
    \centering
    \includegraphics[width=\textwidth]{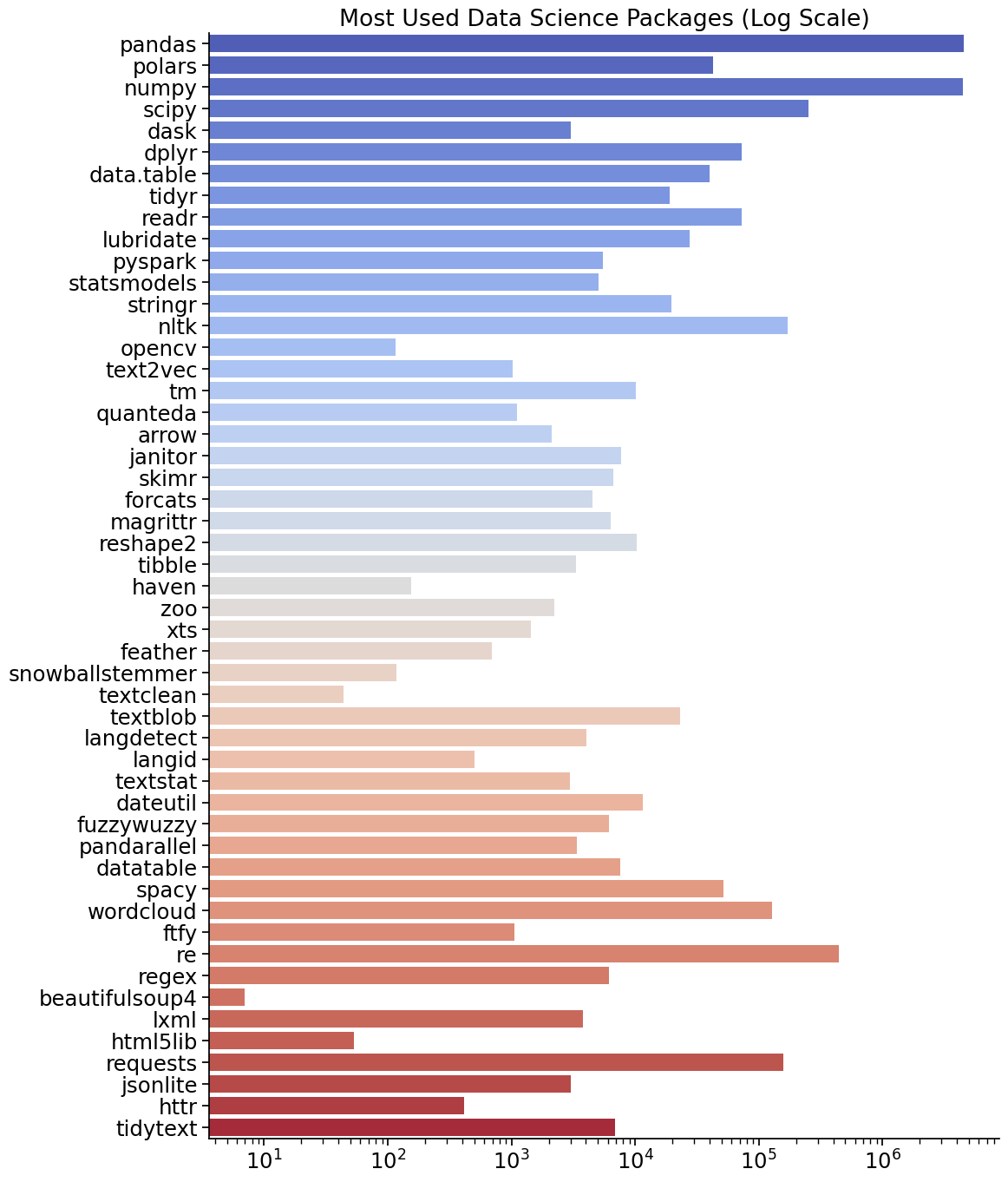}
  \end{subfigure}

  \vspace{0.2em} 

  \begin{subfigure}[b]{0.49\textwidth}
    \centering
    \includegraphics[width=\textwidth]{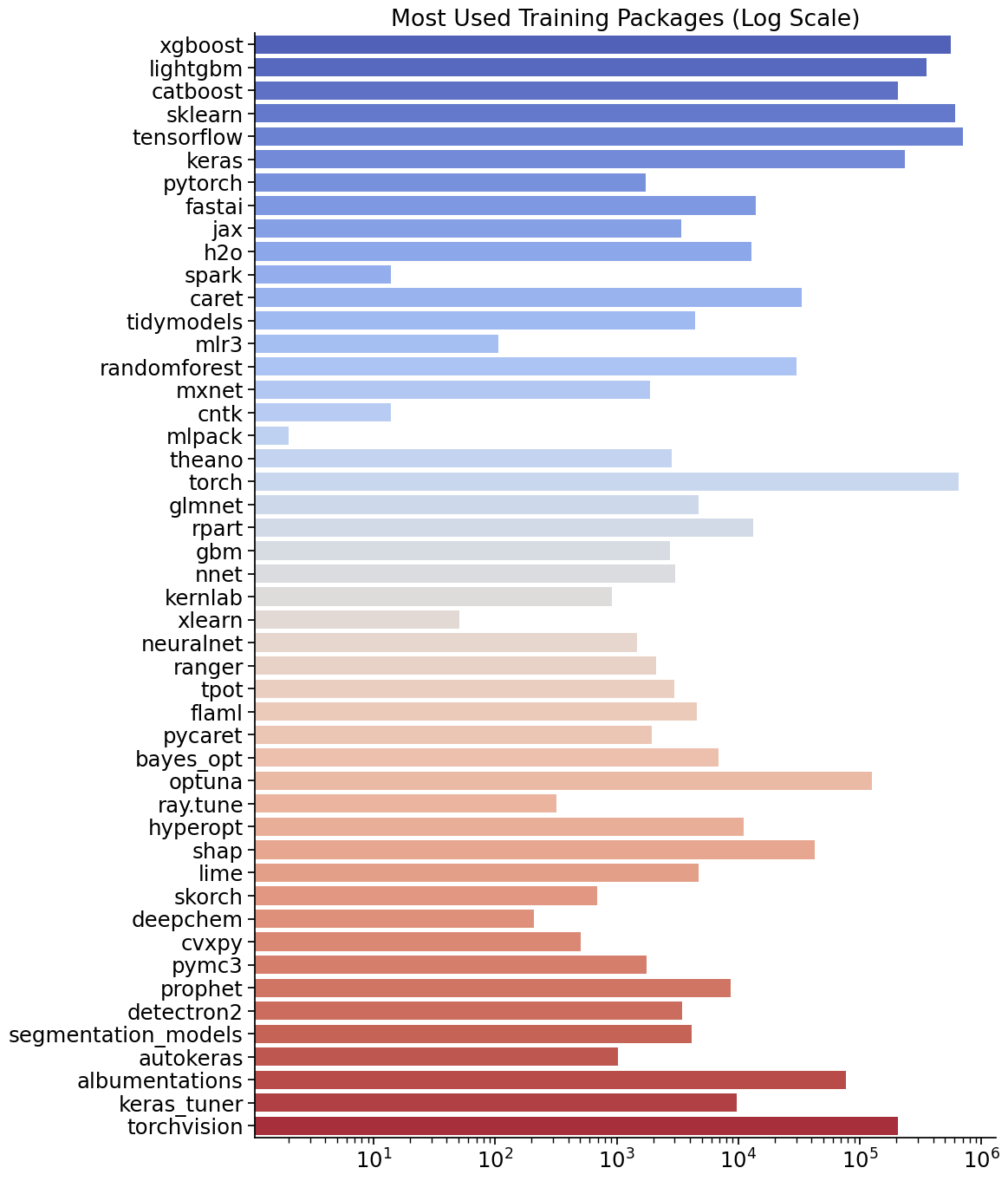}
  \end{subfigure}
  \hfill
  \begin{subfigure}[b]{0.49\textwidth}
    \centering
    \includegraphics[width=\textwidth]{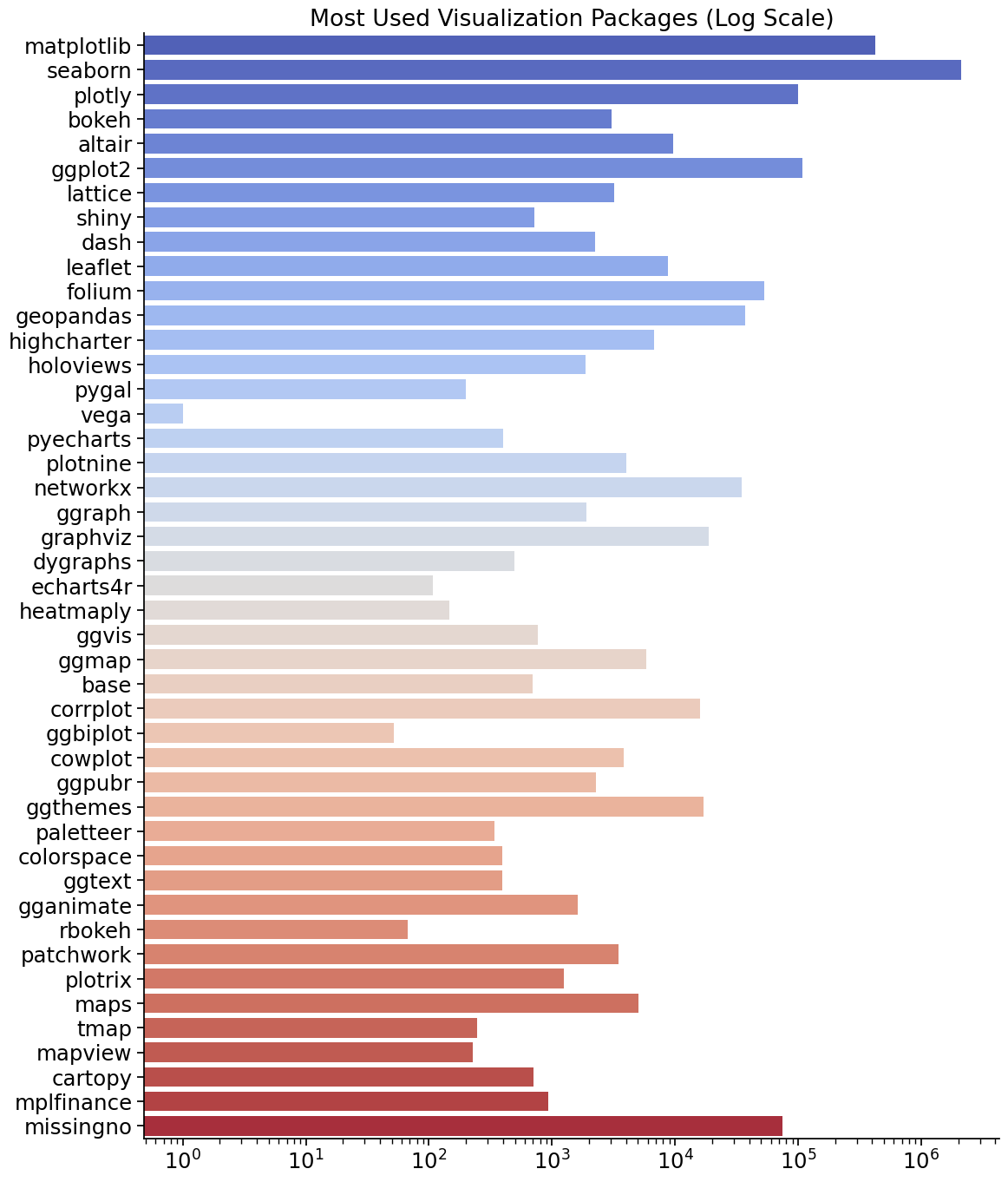}
  \end{subfigure}

  \caption{Distributions of imported packages across four categories in Kaggle kernels: Visualization, Training, Data Science, and ML/LLM Models.
  The x-axis represents the total imports, displayed on a logarithmic scale \kcite{notebook}{https://www.kaggle.com/code/kevinbnisch/most-used-packages-methods-on-kaggle}.
}
  \label{fig:package-categories}

\end{figure}

To obtain a more fine-grained view of the broader range of packages used on Kaggle, we categorized all packages into four groups: \textbf{Visualization}, \textbf{Training}, \textbf{Data Science}, and \textbf{ML/LLM Models}.  
To construct these categories, we compiled lists of commonly used packages for each category by consulting various online sources (e.g., Medium articles, Kaggle discussions) and by querying large language models such as ChatGPT and Gemini to further expand the lists \kcite{notebook}{https://www.kaggle.com/code/kevinbnisch/most-used-packages-methods-on-kaggle}.  
For each category, we assembled a set of 50 representative packages and then assigned package imports from Kaggle kernels to the appropriate category based on these lists, as outlined by Figure \ref{fig:package-categories}.

The x-axis of all four charts uses a logarithmic scale to represent the total number of imports across all kernels.  
As such, bars that appear to be of similar length may still differ by several orders of magnitude in actual import frequency.

For \textbf{visualization}, \texttt{seaborn} seemingly leads the way, followed closely by \texttt{matplotlib} and \texttt{plotly}--the core plotting tools in the Python data science stack.  
However, this instance is somewhat skewed: since we consider only base imports, it excludes submodules such as \texttt{matplotlib.pyplot}, which--as shown in Figure~\ref{fig:side_by_side}--is used more frequently than \texttt{seaborn}. This should be kept in mind when interpreting the results.
Nevertheless, each can serve a distinct role: \texttt{matplotlib} for low-level control, \texttt{seaborn} for statistical plots, and \texttt{plotly} for interactive visuals.

Other Python libraries like \texttt{bokeh}, \texttt{altair}, and \texttt{holoviews} appear less often, likely due to being less known or specialized use cases.
R packages such as \texttt{ggplot2}, \texttt{lattice}, and \texttt{shiny} are present but much less common, as to be expected given Python's dominance.

Geospatial libraries--including \texttt{folium}, \texttt{leaflet}, \texttt{geopandas}, \texttt{cartopy}, and \texttt{mapview}--show up in the long tail, mainly tied to location-based tasks.
Packages like \texttt{missingno} and \texttt{heatmaply} highlight how visualization also supports data quality checks and exploration.  
This might help explain why traditional cleaning methods appear less frequently—many preprocessing steps happen through such specialized tools with custom method calls.

For \textbf{training}, the top spots go to the familiar trio: \texttt{xgboost}, \texttt{catboost}, and \texttt{lightgbm}, probably favored by kagglers for their speed, accuracy, and ease of use with tabular data.  
They’re followed by \texttt{sklearn} for classical ML and \texttt{tensorflow}, which--along with \texttt{keras} and \texttt{pytorch}--has long been a core deep learning framework.  
The strong presence of both \texttt{keras} and \texttt{pytorch} shows that while gradient boosting is still dominant, neural networks are just as popular, especially for image, text, or time series tasks.  
Newer tools like \texttt{jax}, \texttt{fastai}, and \texttt{torchvision} are gaining ground, though they still trail the main libraries.

In the mid-range, R packages like \texttt{caret}, \texttt{mlr3}, and \texttt{randomForest} appear alongside Python tools such as \texttt{h2o}, \texttt{spark}, and \texttt{tidymodels}--suggesting the obvious: Python clearly leads in model training.

A long tail of specialized tools--including \texttt{optuna}, \texttt{ray.tune}, \texttt{hyperopt}, and \texttt{flaml}--points to growing interest in AutoML and hyperparameter tuning, especially in competitive settings (see Section~\ref{ssec:comp-trends}).

For \textbf{data science}, the top libraries once again reflect the core Python stack: \texttt{pandas} and \texttt{numpy} remain essential for data manipulation and numerical computing, with the addition of \texttt{scipy}, providing fundamental algorithms.  
The strong showing of \texttt{polars}, a faster alternative to \texttt{pandas}, suggests more interest in high-performance tools, possibly due to the rise of larger datasets.  
Further down, libraries like \texttt{dask}, \texttt{statsmodels}, and \texttt{pyspark} point to the need for scalable and statistical workflows. On the R side, \texttt{data.table}, \texttt{tidyr}, \texttt{dplyr}, and \texttt{readr} reflect continued use of tidy data principles.  

Many mid-ranked packages are tied to natural language processing--including \texttt{nltk}, \texttt{text2vec}, \texttt{textblob}, \texttt{spacy}, \texttt{langdetect}, and \texttt{wordcloud}--highlighting the trend of text analysis. The presence of \texttt{re} and \texttt{regex} underlines the ongoing role of regular expressions in preprocessing.
Tools like \texttt{janitor}, \texttt{skimr}, \texttt{forcats}, and \texttt{reshape2} stress the importance of data cleaning, while packages such as \texttt{fuzzywuzzy} and \texttt{pandarallel} address string matching and parallel processing.
Finally, libraries like \texttt{html5lib}, \texttt{beautifulsoup4}, \texttt{requests}, and \texttt{httr} suggest that web scraping and API-based data collection are somewhat common in Kaggle workflows.

For \textbf{ML--LLM Models}, we see a blend of classic machine learning and modern transformer-based approaches. Leading the chart by a lot (remember the log scale) is \texttt{transformers}, the go-to Hugging Face library for working with pretrained language models--showcasing the growing role of LLMs in applied workflows, even within Kaggle notebooks, which was to be expected.  
Close behind are \texttt{bert} and \texttt{xgbclassifier}, representing two key paradigms: transformers for text-heavy tasks, and gradient boosting for structured data. \texttt{randomforestclassifier} also ranks high, continuing the popularity of ensemble methods in traditional ML.

There is also growing interest in newer LLMs such as \texttt{llama}, \texttt{t5}, \texttt{gpt2}, and \texttt{deberta}.  
Mentions of models like \texttt{gemma}, \texttt{bloom}, \texttt{flan}, and \texttt{claude} suggest that open-weight models are being adopted soon after their release, although their usage remains limited--likely due to hardware or platform constraints, but also because of their novelty in comparison to other technologies.

Classic \texttt{scikit-learn} models such as \texttt{logisticregression}, \texttt{svm}, \texttt{knn}, and \texttt{naivebayes} seem to remain steady, possibly valued for their simplicity and interpretability. Tree-based models like \texttt{decisiontree}, \texttt{baggingclassifier}, and \texttt{stackingclassifier} also appear, though they’re less prominent than ensembles like \texttt{xgbclassifier}.
The presence of tools like \texttt{autogluon}, \texttt{huggingface}, and \texttt{sagemaker} points to a shift toward more automated or cloud-friendly workflows that streamline model development and deployment.

\subsubsection{\texttt{torch} vs \texttt{tensorflow}}

\begin{figure}[htpb]
    \centering
    \includegraphics[width=\linewidth]{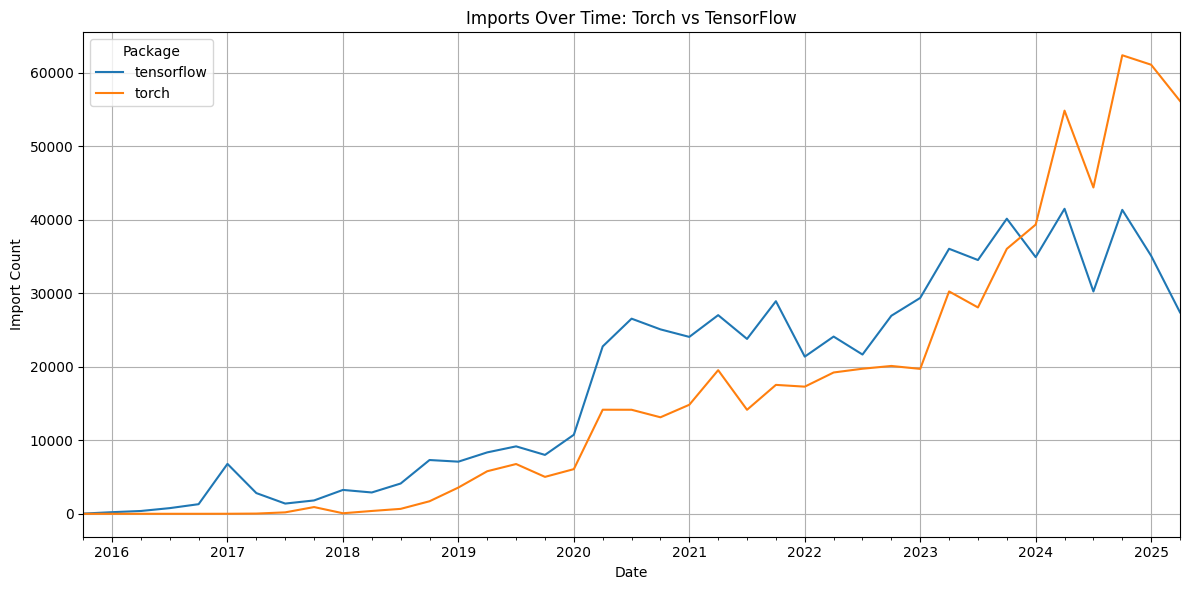}
    \caption{\texttt{torch} vs \texttt{tensorflow} over time. We can see a rise in \texttt{torch} users \kcite{notebook}{https://www.kaggle.com/code/kevinbnisch/most-used-packages-methods-on-kaggle}.}
    \label{fig:torch-vs-tensorflow}
\end{figure}

The difference in usage between \texttt{torch} and \texttt{tensorflow} caught our attention, as the latter appears to dominate the former in overall import counts. 
However, based on our own experience and broader trends within the machine learning community, we have observed a strong shift toward \texttt{torch} in recent years.  
When plotting the usage of both libraries over time, this intuition is confirmed: \texttt{torch} has been overtaking \texttt{tensorflow} since late 2023 (see Figure \ref{fig:torch-vs-tensorflow}).

\subsection{Competition Trends \& Insights} \label{ssec:comp-trends}

\vspace{-0.5cm}
\tldr{
    First, we examine the evolution of imported packages in competition kernels: \texttt{xgboost} dominated early years, later replaced by \texttt{lightgbm}, \texttt{tensorflow}, and \texttt{transformers}, reflecting a broader community shift from classical ML to deep learning and NLP. Packages like \texttt{sklearn} remain foundational, while newer entries such as \texttt{autogluon}, \texttt{optuna}, and \texttt{openai} indicate growing interest in AutoML and generative models. Explainability libraries (\texttt{shap}, \texttt{lime}) are also increasingly common.
    \newline
    \newline
    Secondly, we investigate discrepancies between public and private leaderboard scores, outlining consistent gaps that highlight the risks of leaderboard overfitting. 
    Using visualizations and metric-based analyses, we show how real-world generalizability can differ massively from public leaderboard performance--as seen in competitions like \textit{Optiver} and \textit{M5 Forecasting}.
    However, our analysis also shows that, on average, Kagglers produce only a 5--10\% discrepancy between private and public leaderboard scores.
}

Since 2010, Kaggle has been hosting data sience competitions that offer medals and prize money for the best solutions.  
To truly participate in a regular competition, a Kaggler must submit a kernel that produces a specified output, which is then evaluated against the competition's defined metric.  
Thanks to the \textsc{Kaggle Meta Code} dataset, we have access to these competition kernels from 2015 onward, along with their method calls and imported packages (which we extracted in a previous Section \kcite{dataset}{https://www.kaggle.com/datasets/kevinbnisch/kaggles-most-used-packages-and-method-calls}), but also meta data such as public and private leaderboard scores.  

\subsubsection{What Packages are Kagglers using in Competitions?}
 
\begin{figure}
    \centering
    \includegraphics[width=0.9\linewidth]{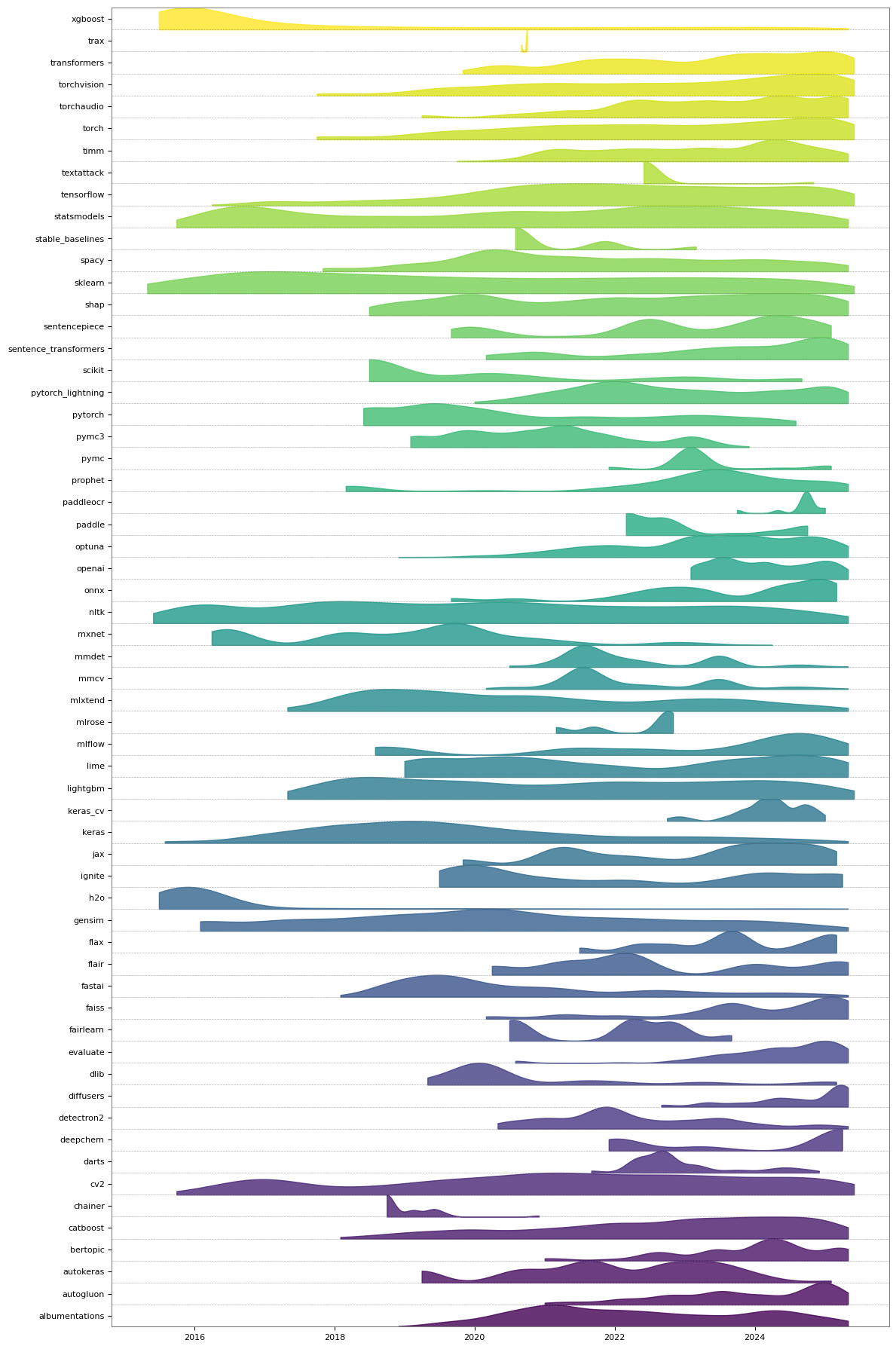}
    \caption{Ridge plots illustrating packages used in competitions over time \kcite{notebook}{https://www.kaggle.com/code/kevinbnisch/what-packages-are-kagglers-using-in-competitions/notebook}.}
    \label{fig:comp-packages-over-time}
\end{figure}

As a first step, we analyze the evolution of imported packages in competition kernels over time \kcite{notebook}{https://www.kaggle.com/code/kevinbnisch/what-packages-are-kagglers-using-in-competitions/notebook}.  
To do this, we first normalize all sub-imports to their base namespace (e.g., \texttt{xgboost.core} becomes \texttt{xgboost}). We then filter out non-essential packages--that is, packages not directly related to model building, prediction, or competitive data science \kcite{dataset}{https://www.kaggle.com/datasets/kevinbnisch/ml-and-ai-related-imports} (e.g., \texttt{os}, \texttt{warnings}) and plot the results.  
Figure~\ref{fig:comp-packages-over-time} presents each major technology and its usage in competition kernels over time, visualized as ridge plots.

A few clear trends emerge. Notably, the usage of \texttt{xgboost} was dominant in the early years (2015--2017), reflecting the popularity of gradient boosting methods in structured data competitions during that period. However, its usage tapered off significantly after 2018, giving way to deep learning frameworks, but also similar techniques such as \texttt{lightgbm}.
The plot even indicates a seamless swap from \texttt{xgboost} to \texttt{lightgbm}.

The rise of \texttt{tensorflow}, \texttt{pytorch}, and \texttt{transformers} is evident starting around 2018, with especially strong growth for \texttt{transformers} from 2020 onward. This aligns with the widespread adoption of transformer-based models (such as BERT and GPT) in both research and applied machine learning tasks. The HuggingFace ecosystem, including \texttt{datasets}, \texttt{tokenizers}, and \texttt{textattack}, shows increasing relevance, emphasizing the growing focus on NLP competitions.
\texttt{sentencepiece} and \texttt{sentence-transformers} show usage spikes coinciding with the adoption of transformer models and multilingual NLP pipelines, which fits the overall transformer-based narrative.

Libraries like \texttt{scikit-learn} (\texttt{sklearn}) maintain a steady baseline, highlighting their foundational role in preprocessing, model evaluation, and classical ML methods. 
Interestingly, \texttt{shap} and \texttt{lime} show noticeable presence in recent years, suggesting a rising awareness around explainability in model development, even in competitive settings.

In recent years (2022–2025), newer frameworks such as \texttt{openai}, \texttt{diffusers}, and especially \texttt{autogluon} have emerged, indicating a strong shift towards generative models and AutoML pipelines. The increasing presence of \texttt{optuna} points to a growing focus on automated hyperparameter optimization.
Conversely, packages like \texttt{mxnet}, \texttt{chainer}, and \texttt{h2o} have largely diminished, indicative of a transition in community preference and tool support. Similarly, \texttt{fastai} shows a peak followed by a decline.

Finally, the long tail of niche or specialized packages (e.g., \texttt{faiss}, \texttt{cv2}, \texttt{mmcv}, \texttt{darts}, and \texttt{evaluate}) highlights the variety of tools employed in specific types of competitions such as computer vision, time series, or hardware-constrained tasks.

Overall, the evolution of Kaggle competitions show a shift from tree-based methods to deep learning and transformer-centric approaches, with increasing attention to interpretability, reproducibility, and automation.

\subsubsection{Public vs. Private Leaderboard Scores}

In a typical Kaggle competition, both a public and a private leaderboard exist.  
The public leaderboard is, as the name suggests, publicly visible and displays the rankings of participants based on their submitted kernels.  
Each Kaggler is represented on the public leaderboard by the submission with their \textbf{highest public leaderboard} score.  

What remains hidden during the competition is the private leaderboard.  
This leaderboard is computed using a different test set than the one used for the public leaderboard. For every submission evaluated, a corresponding \textbf{private leaderboard score} is calculated as well--but not shown.  
After the competition concludes, the private leaderboard is revealed, and it determines the final rankings and official winners.  
This means that a participant who ranks first on the public leaderboard for an entire month could theoretically end up far lower--or even last--once the private leaderboard is revealed.

Kaggle implements this mechanism to mitigate overfitting to a single public leaderboard dataset.  
When participants fine-tune their models purely to improve public leaderboard performance, they risk overfitting the specific test set used for public evaluation. This reduces the generalizability and real-world applicability of the model.  
The private leaderboard thus acts as a safeguard to ensure that winning models perform well on unseen data. In this sense, the public leaderboard is merely an indication--not the final outcome.
Fortunately, we have access to both the public and private leaderboard scores for all submitted competition kernels.

\begin{figure}
    \centering
    \includegraphics[width=0.95\linewidth]{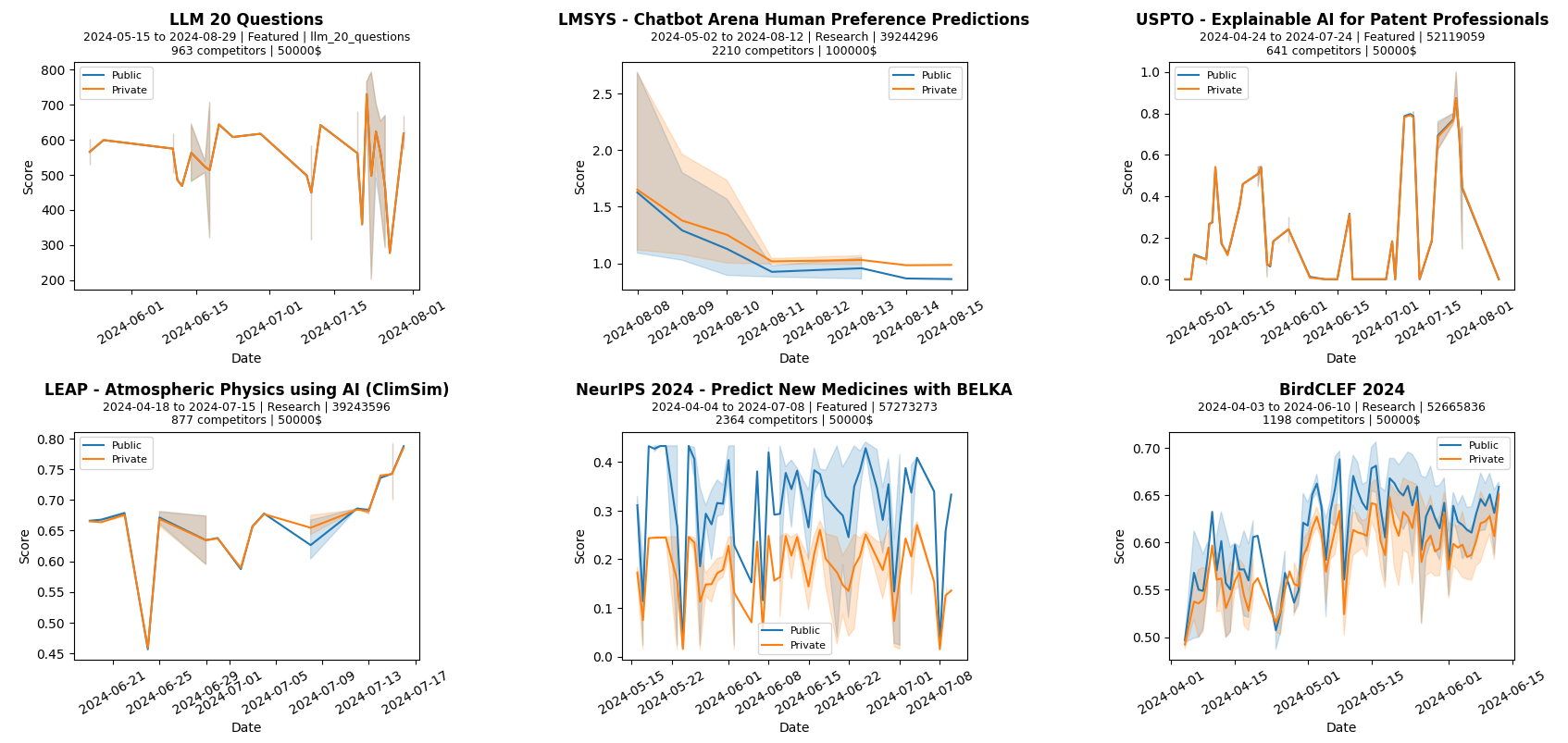}
    \caption{Recent six competitions and their private vs. public leaderboard scoring \kcite{notebook}{https://www.kaggle.com/code/kevinbnisch/what-packages-are-kagglers-using-in-competitions/notebook}.}
    
    \vspace{0.3cm}
    
    \includegraphics[width=0.95\linewidth]{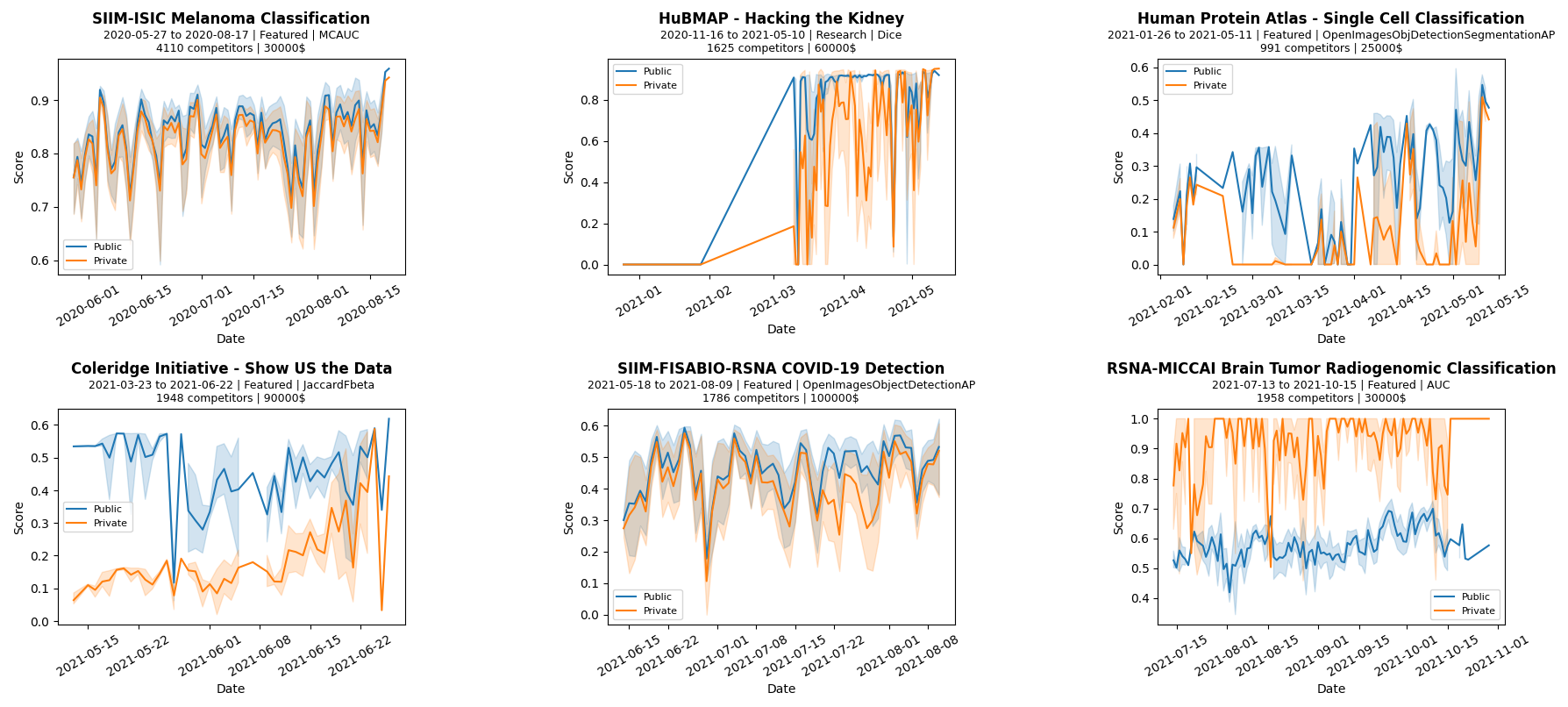}
    \caption{Competitions with biggest average private/public leaderboard score difference \kcite{notebook}{https://www.kaggle.com/code/kevinbnisch/what-packages-are-kagglers-using-in-competitions/notebook}.}
    
    \label{fig:leaderboard-diff}
\end{figure}

Figure~\ref{fig:leaderboard-diff} displays a selection of competitions (for more competitions, refer to \kcite{notebook}{https://www.kaggle.com/code/kevinbnisch/what-packages-are-kagglers-using-in-competitions/notebook}) along with the evolution of public and private leaderboard scores over time.  
The top row shows recent competitions, sorted by date, while the bottom row highlights competitions with the highest average discrepancy between public and private leaderboard scores throughout the duration of the competition.  
In an ideal scenario, the public and private leaderboard scores should align closely, indicating that submitted models generalize well across both evaluation sets.  

Generally, public and private leaderboard scores tend to evolve in a parallel fashion over time, which is desirable, although both can exhibit considerable variance.  
In most cases, the two scores are not identical, and in some extreme instances--such as the \textit{RSNA-MICCAI Brain Tumor} competition--they operate on entirely different scales, indicating substantial divergence between the public and private evaluation sets.

The plots also show numerous spikes in both score variance and average progression over time.  
Since the visualizations include all submissions from all teams across the full duration of each competition, such fluctuations are expected.  
They highlight the wide range of modeling approaches employed--some more successful than others--and show the experimental nature of many competition strategies.

The key takeaway, of course, is that optimizing for a single leaderboard--particularly the public one--does not guarantee good generalizability or robust modeling performance.  
These plots, and Kaggle competitions more broadly, illustrate this phenomenon clearly.

\begin{figure}
  \centering
  \begin{minipage}[b]{0.49\textwidth}
    \centering
    \includegraphics[width=\textwidth]{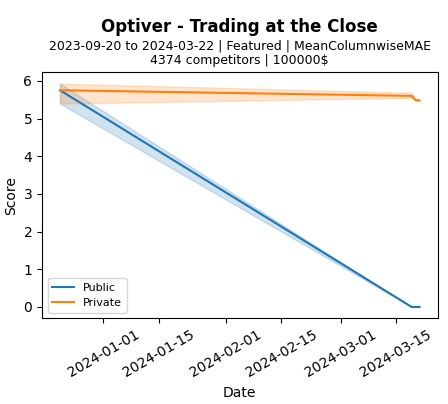}
  \end{minipage}
  \hfill
  \begin{minipage}[b]{0.49\textwidth}
    \centering
    \includegraphics[width=\textwidth]{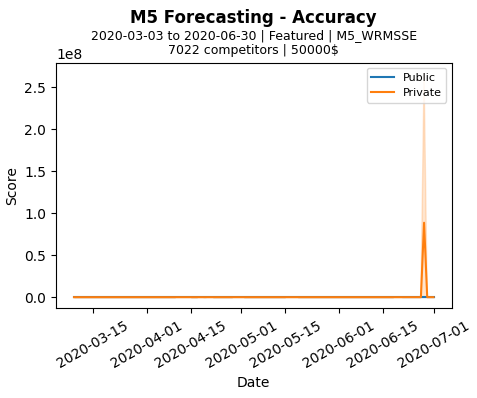}
  \end{minipage}
  \caption{Two exemplary competitions, on the left the \textit{Optiver} competition and on the right the \textit{M5 Forecasting} competition \kcite{notebook}{https://www.kaggle.com/code/kevinbnisch/what-packages-are-kagglers-using-in-competitions/notebook}.}
  \label{fig:optiver}
\end{figure}

We now take a closer look at two specific competitions outlined in Figure \ref{fig:optiver}: the \textit{Optiver} and \textit{M5 Forecasting} competitions.

The \textit{Optiver} competition serves as a good example of the contrast between real-world modeling and controlled test scenarios.
In this competition, lower scores indicate better performance according to the official evaluation metric.  
As shown in the plot, public leaderboard scores steadily decrease over time, approaching near-perfect values. 
In contrast, private leaderboard scores show more variance and only a very gradual improvement.

In the competition, the timeline was divided into two distinct phases: a training phase and a forecasting phase.  
During the training phase, participants were allowed to train, test, and submit their models as usual. Once this phase concluded, no further submissions were accepted.  
In the subsequent forecasting phase, real-world data was used to evaluate the submitted models by having them predict the closing prices of hundreds of Nasdaq-listed stocks.  
All results from this phase were reflected exclusively in the private leaderboard scores.
As a result, the private leaderboard scores more accurately reflect the true capabilities of the models, while the public leaderboard can be easily adapted to--and overfitted--highlighting the risks of relying on static test sets and controlled evaluation scenarios.

Furthermore, the \textit{M5 Forecasting} competition is a classic case of drastic public vs. private leaderboard shifts. 
Participants forecasted 28 daily Walmart item sales across two phases: a public phase with live leaderboard feedback, and a private phase revealed only after the deadline.

Many top teams tuned models to short-term trends seen in the public period, notably applying \enquote{magic multipliers} (e.g., $\times$1.02–1.05) to inflate forecasts, assuming June sales would rise as in prior years. 
However, the expected trend broke in June 2016, causing such models to overshoot and rank poorly on the private leaderboard.

This led to a massive shake-up: some public top-100 teams plummeted, while others who avoided these tweaks rose sharply (\numprint{1000} - \numprint{5000} places up). 
Robust models relying on cross-validation without magic numbers generalized better, showing that leaderboard-driven tuning can heavily backfire when trends shift.
To summarize in the \href{https://www.christophenicault.com/post/m5_forecasting_accuracy/#:~:text=There%20was%20a%20huge%20shake,working%20only%20for%20that%20period}{words of Christophe Nicault}, a participant of the competition:

\textit{\enquote{There was a huge shake-up for this competition with different results between the public leaderboard and the private leaderboard. Part of it can be explained by the difference for this period compare to the previous years. For the 5 previous year, the sales in June have always increased compared to the sales in May. The trend was up during the previous month, and most of the model were being optimistic on the volume of sales. For the validation period, many teams used a multiplier coefficient to reflect the trend in the forecast, leading to a better score for the public data, but working only for that period.}}
\newline
\newline
\textbf{Average Public/Private Score Discrepancy Over Time}

\begin{figure}
    \centering
    \includegraphics[width=\linewidth]{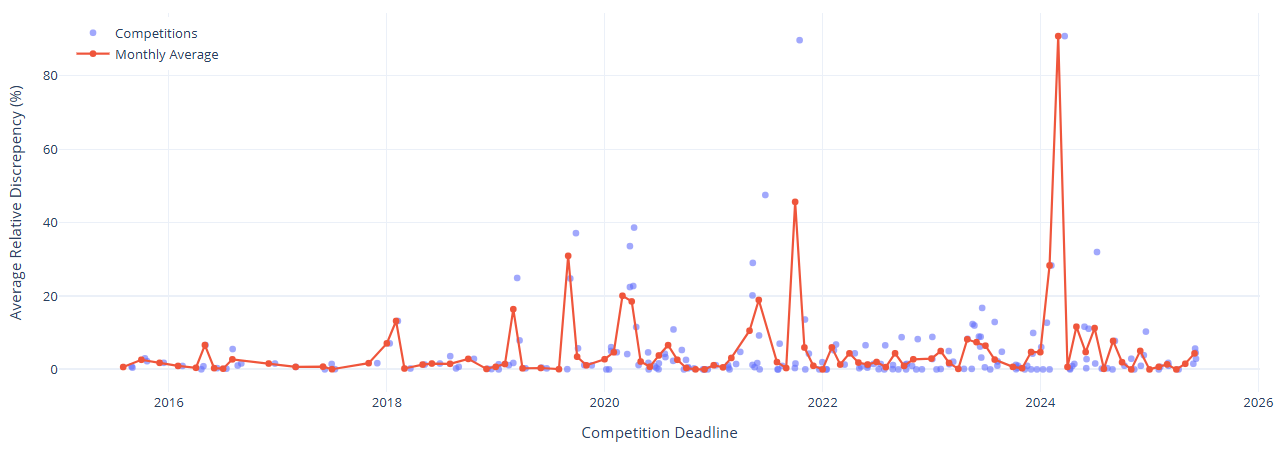}
    \caption{Showcases the average normalized discrepancy percentage between public and private leaderboard scores over time \kcite{notebook}{https://www.kaggle.com/code/kevinbnisch/what-packages-are-kagglers-using-in-competitions/notebook}.}
    \label{fig:avg-discrepancy}
\end{figure}

We are also intersted in the \textit{average normalized discrepancy percentage} between public and private leaderboards across all Kaggle competitions over time.
So for each competition, we compute the average \textit{relative change} between public and private scores for every participant, normalized by the range of public scores and outline it through Figure \ref{fig:avg-discrepancy}. The formula is:

\begin{equation}
\text{avg\_rel\_diff\_pct} = \left( \frac{1}{N} \sum_{i=1}^{N} \left| \frac{\text{public}_i - \text{private}_i}{\max(\text{public}) - \min(\text{public})} \right| \right) \cdot 100
\end{equation}

where:

\begin{align*}
N & = \text{number of participants}, \\
\text{public}_i & = \text{public score of participant } i, \\
\text{private}_i & = \text{private score of participant } i
\end{align*}

Each value (purple dots) reflects the \textit{average percentage discrepancy} per competition. For example, a value of 15\% means that on average, each participant's score changed by 15\% of the public leaderboard range when switching to the private leaderboard.
Finally, we aggregate these per-competition discrepancy values by the month of the competition deadline, to observe how leaderboard volatility has evolved over time (red line).

The high spike observed in early 2024 is attributed to the aforementioned \textit{Optiver} competition, which exhibited a significant discrepancy between public and private leaderboard scores.  

A smaller spike, occurring at the end of 2021, is primarily due to the \textit{RSNA-MICCAI Brain Tumor Radiogenomic Classification} competition (indicated by the purple dot at the top).  
In this competition, participants were tasked with predicting the genetic subtype of glioblastoma using MRI (magnetic resonance imaging) scans to detect the presence of MGMT promoter methylation.  
However, the challenge was hindered by the limited number of available samples--approximately 310--which resulted in leaderboard scores that were nearly indistinguishable from random submissions.  
As a consequence, no strong models could be developed, leading to an unusually large gap between public and private leaderboard performance.

Aside from these anomalies, the graph shows an overall average discrepancy of approximately 5--10\% between private and public submission scores, which shows the importance of real-world evaluation and testing.  
Nevertheless, within this context, Kaggle participants generally do a commendable job of minimizing this gap.

\subsection{What Technologies do (winning) Kagglers use?}
\vspace{-0.5cm}
\tldr{
    We extract technologies from \numprint{4419} solution write-ups using LLM-based parsing, showing that winning Kaggle solutions most frequently mention \texttt{efficientnet}, \texttt{lightgbm}, \texttt{data augmentation}, and ensemble methods. 
    Both classical and deep learning approaches co-exist, with strong emphasis on training techniques such as \texttt{cross-validation}, \texttt{label smoothing}, and \texttt{mixup}.
    We also demonstrate high entropy and technological diversity in competition write-ups--both per competition and per year--with a steady increase observed over time.
}

After a competition concludes, Kaggle participants are encouraged to publish a \textit{solution write-up}--a forum post in which they explain their approach, the technologies used, and any noteworthy modeling decisions.  
This practice is especially encouraged for top-performing teams, who are often \href{https://www.kaggle.com/discussions/general/427114}{incentivized with additional monetary rewards or recognition}.  
This serves as another highly valuable resource for analyzing the technological landscape within the competition ecosystem.
Consequently, we collected all available solution write-ups from the competition discussion sections and analyzed them \kcite{notebook}{https://www.kaggle.com/code/kevinbnisch/what-technologies-do-winning-kagglers-use}.  
We used a large language model (specifically, \texttt{gpt-4o-min}) to extract all mentions of technologies from the full text and structure them into a list.  
In addition, we prompted the model to generate concise summaries of each write-up.  
To group and standardize different references to the same technology (e.g., \texttt{AdamW}, \texttt{AdamW Optimizer}, or \texttt{AdamW Optim}), we applied a simple fuzzy-matching algorithm to group equivalent mentions.  
As a result, we obtained a structured dataset consisting of \numprint{4419} competition write-ups--primarily from top-ranking and winning teams--along with extracted technologies, techniques, and generated summaries. 
This dataset has been made publicly available on Kaggle \kcite{dataset}{https://www.kaggle.com/datasets/kevinbnisch/competition-writeups-with-technologies-and-summary}.

\begin{figure}
    \centering
    \includegraphics[width=0.9\linewidth]{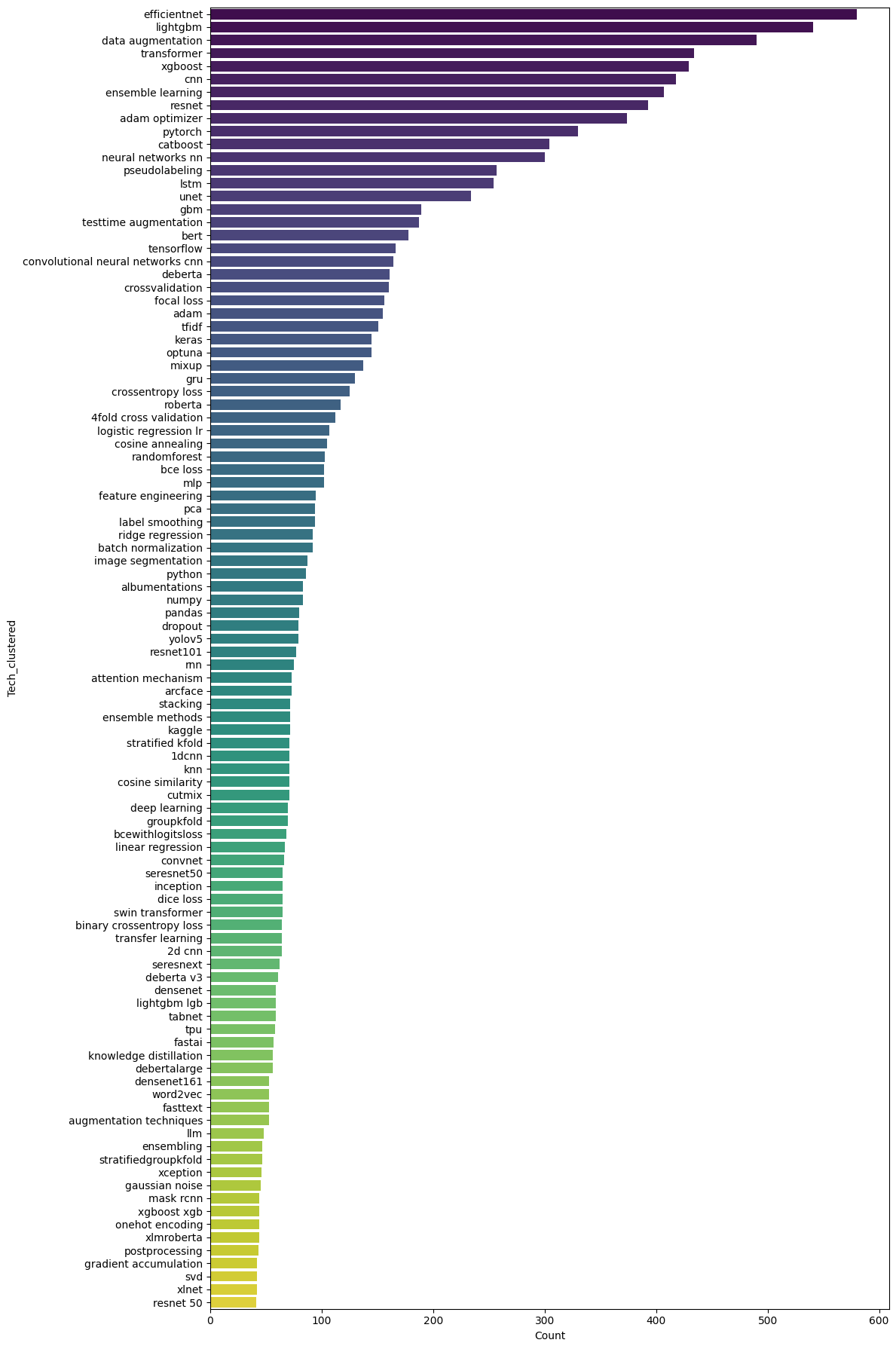}
    \caption{The top technologies mentioned in solution write-ups \kcite{notebook}{https://www.kaggle.com/code/kevinbnisch/what-technologies-do-winning-kagglers-use}.}
    \label{fig:technologies-writeups}
\end{figure}

Figure~\ref{fig:technologies-writeups} displays the top technologies mentioned in Kaggle solution write-ups.  
It is important to note that the mere mention of a technology does not necessarily indicate that it was a key contributor to the team's success.  
In many cases, teams also describe methods and approaches that did not perform well, although the primary focus of most write-ups is typically on what ultimately worked.

Firstly, the plot reveals a clear dominance of widely adopted machine learning frameworks and techniques.  
\texttt{efficientnet}, \texttt{lightgbm}, and \texttt{data augmentation} stand out as the most frequently referenced tools. This is particularly noteworthy in the case of \texttt{efficientnet}, which was only introduced by Google in 2019, yet has rapidly become a staple in many--supposedly winning--solutions.  
The prominence of \texttt{data augmentation} is also significant, as it reflects a focus on generating additional training data beyond the fixed public dataset.  
This suggests that enhancing model robustness through data augmentation remains a highly effective and commonly adopted approach in competitive settings.

Ensemble-based methods such as \texttt{xgboost}, \texttt{cnn}, and general \texttt{ensemble learning} also appear prominently, highlighting a preference for robust model combination approaches.

Interestingly, both classical and deep learning methods are well represented. Traditional models like \texttt{logistic regression}, \texttt{randomforest}, and \texttt{svm} appear alongside deep learning frameworks such as \texttt{pytorch}, \texttt{tensorflow}, and architectures like \texttt{resnet} and \texttt{bert}. This suggests that Kaggle practitioners often blend modern and traditional approaches depending on the nature of the task.

Furthermore, a variety of optimization strategies and training techniques such as the \texttt{adam optimizer}, \texttt{cross-validation}, \texttt{label smoothing}, and \texttt{mixup} are mentioned, highlighting the importance of fine-tuning and robust evaluation in achieving competitive results. Tools used for experimentation and tuning, like \texttt{optuna}, also feature, suggesting an emphasis on hyperparameter optimization.

Lower-ranked entries like \texttt{llm}, \texttt{xception}, and \texttt{gradient accumulation} indicate niche or emerging techniques that may either be less suitable for typical Kaggle problems or are still gaining traction among practitioners.

Overall, the plot highlights a wide range of technologies and approaches employed by participants, while also revealing the strong influence of Google-developed tools--particularly the Transformer architecture and \texttt{efficientnet}.

\begin{figure}[H]
    \centering
    \includegraphics[width=\linewidth]{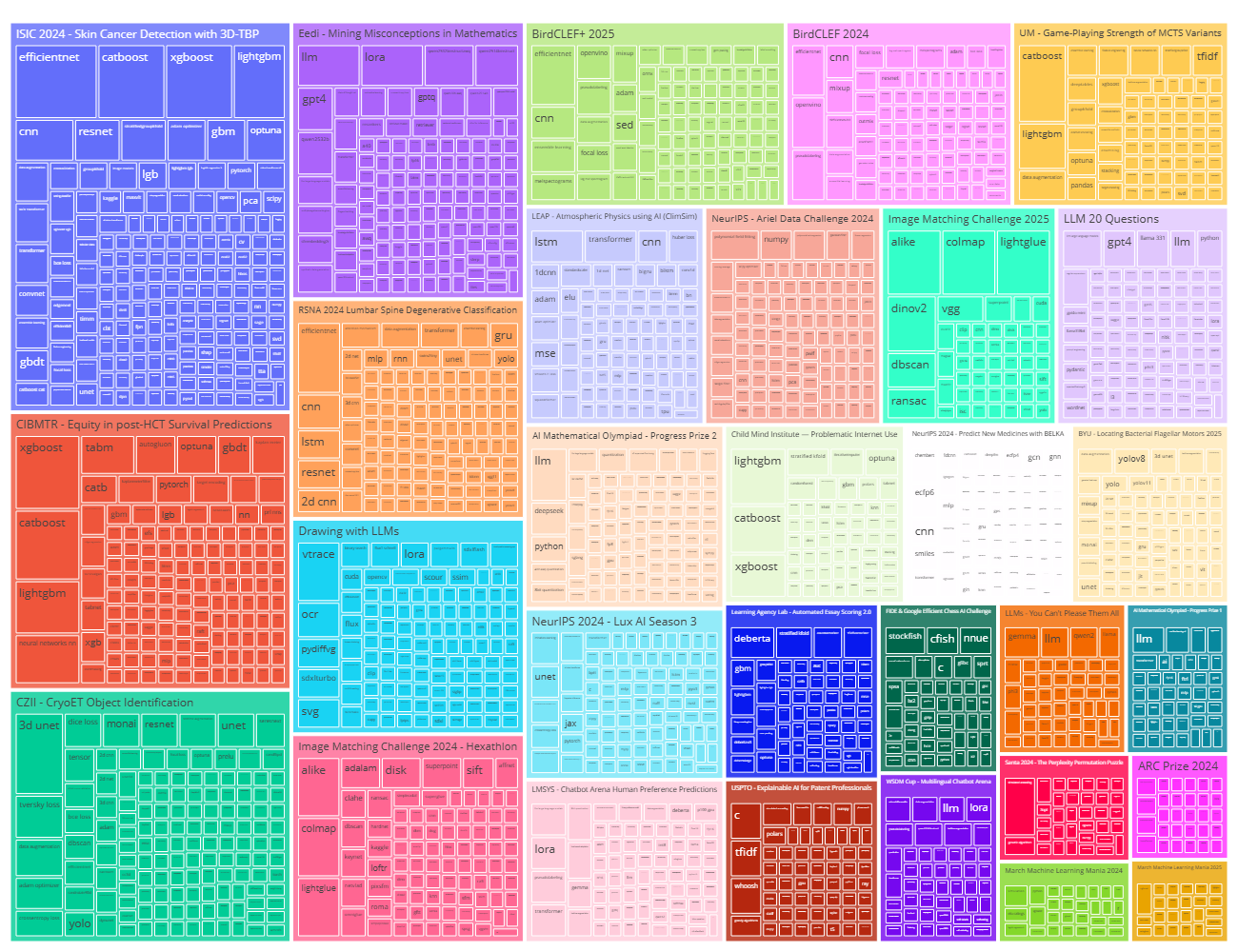}
    \caption{Technologies mentioned within writeups, clustered by their competition and frequency. Refer to \kcite{notebook}{https://www.kaggle.com/code/kevinbnisch/what-technologies-do-winning-kagglers-use} for the full interactive version.}
    \label{fig:tech-writeups-per-comp}
\end{figure}

\begin{figure}[htpb]
    \centering
    \includegraphics[width=\linewidth]{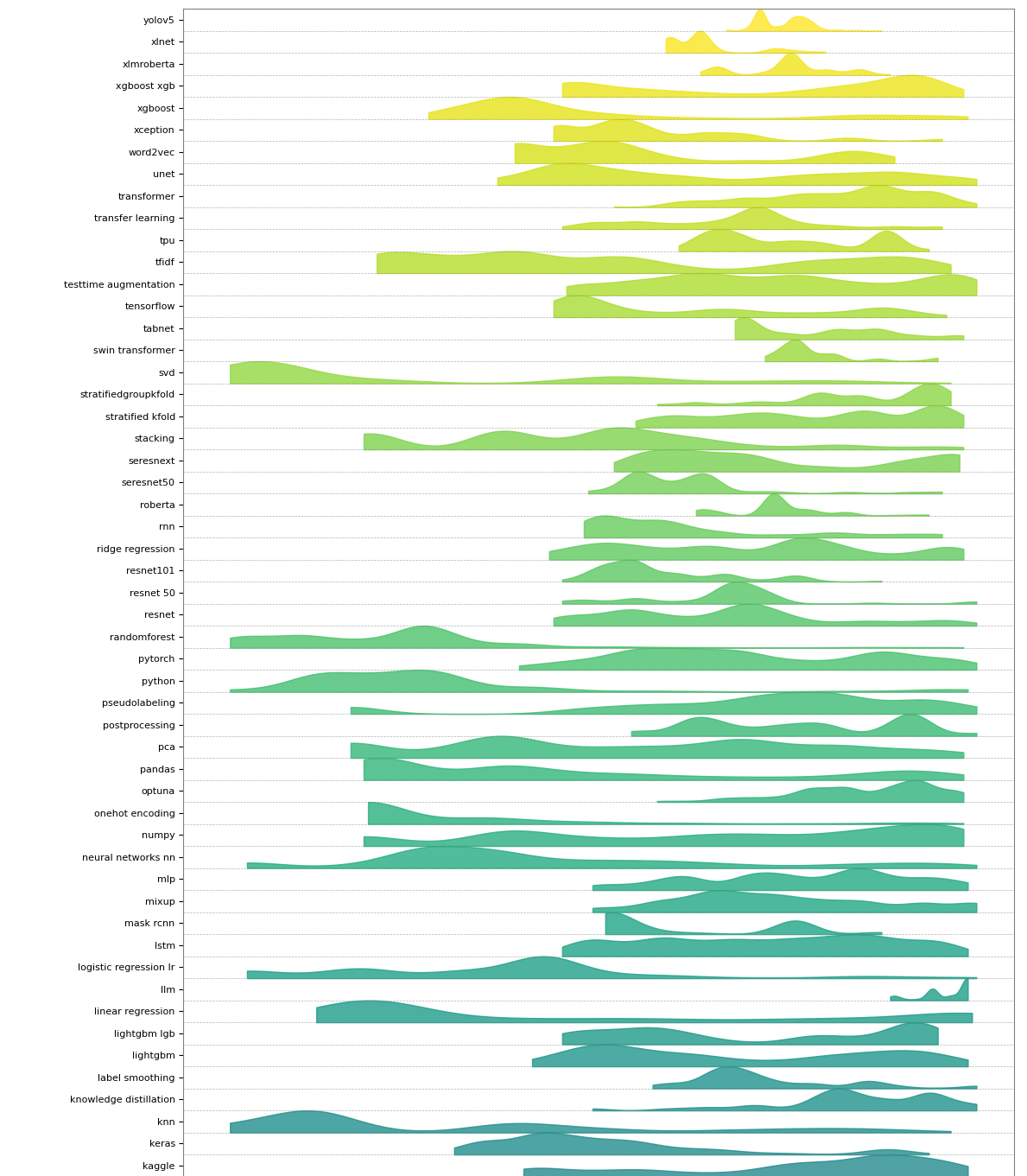}
    \caption{The top technologies mentioned in solution writeups over time (part 1/2) \kcite{notebook}{https://www.kaggle.com/code/kevinbnisch/what-technologies-do-winning-kagglers-use}.}
    \label{fig:top-tech-writeup-time-1}
\end{figure}

\begin{figure}[htpb]
    \centering
    \includegraphics[width=\linewidth]{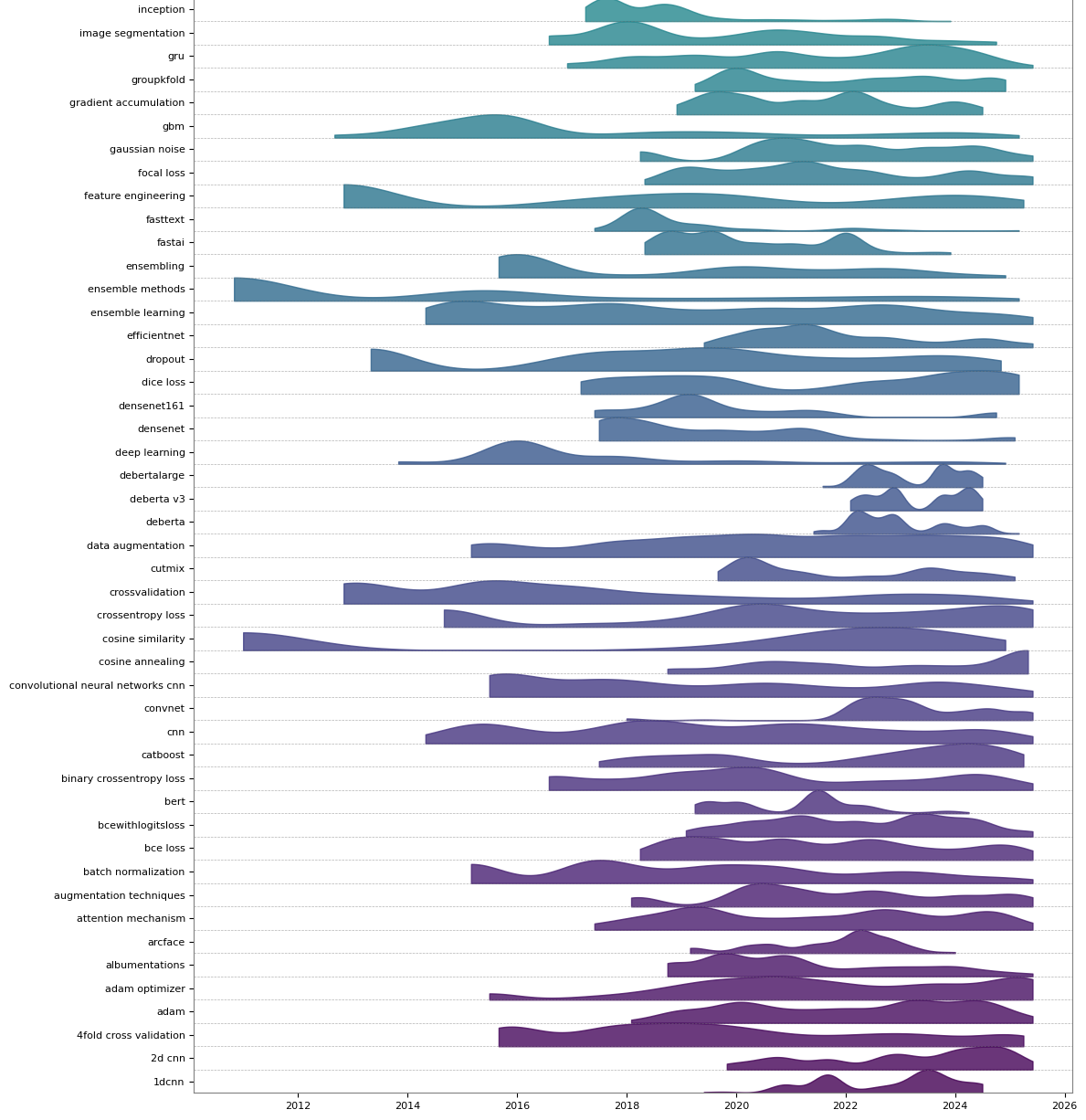}
    \caption{The top technologies mentioned in solution writeups over time (part 2/2) \kcite{notebook}{https://www.kaggle.com/code/kevinbnisch/what-technologies-do-winning-kagglers-use}.}
    \label{fig:top-tech-writeup-time-2}
\end{figure}

In this context, we provide two more visualizations:
\begin{enumerate}
    \item Figures~\ref{fig:top-tech-writeup-time-1} and~\ref{fig:top-tech-writeup-time-2} within the Appendix show the most frequently mentioned technologies in competition write-ups over time, highlighting trends in tool adoption across different years.
    \item Figure~\ref{fig:tech-writeups-per-comp} illustrates the distribution of mentioned technologies clustered by competition and frequency.
\end{enumerate}

\subsubsection{Technology Diversity per Competition over Time}

As we already discussed, each competition typically involves several teams or participants whose write-ups reference various technologies. Herein, the diversity of technologies mentioned can vary significantly:

\begin{itemize}
  \item Some competitions may consistently use only a single dominant technology $\rightarrow$ \textbf{low entropy}
  \item Others may involve a broad and evenly distributed set of technologies $\rightarrow$ \textbf{high entropy}
\end{itemize}

To quantify this, we use \textit{normalized Shannon entropy}, which measures how evenly technologies are distributed across a competition’s write-ups and rescales the entropy to the $[0, 1]$ range for comparability. 

Normalization is necessary because competitions vary in the number of mentioned technologies, the number of participating teams, and the underlying technological landscape, making direct comparisons across competitions otherwise unreliable.

\subsection*{1. Normalized Entropy per Competition}

\begin{figure}[H]
    \centering
    \includegraphics[width=1\linewidth]{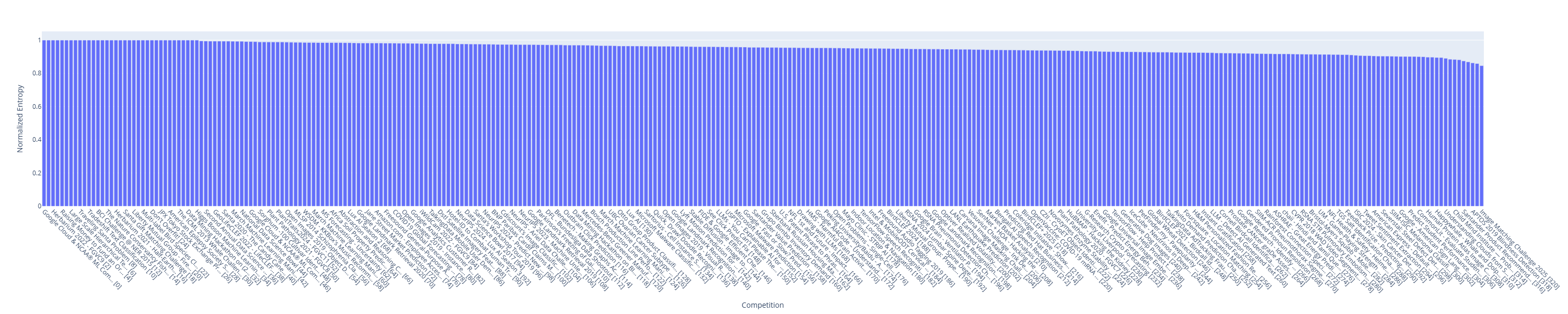}
    \caption{Normalized Entropy per Competition \kcite{notebook}{https://www.kaggle.com/code/kevinbnisch/what-technologies-do-winning-kagglers-use}}
    \label{fig:norm-entropy-per-year}
\end{figure}

Figure \ref{fig:norm-entropy-per-year} shows the \textit{normalized entropy} for each competition launched since 2010. Each bar represents how evenly technologies were mentioned in this competition's write-ups.
In this context, a value close to 0 indicates that a single technology dominated the write-ups for that competition, whereas a value near 1 suggests that multiple technologies were mentioned with similar frequency.  
Here, we only present a zoomed-out, high-level view of the figure to illustrate that the average entropy remains consistently close to 1 (for the complete interactive chart, refer to \kcite{notebook}{https://www.kaggle.com/code/kevinbnisch/what-technologies-do-winning-kagglers-use}).  
This indicates that within competition write-ups, no single technology overwhelmingly dominates. Instead, a general diversity of tools and approaches is consistently present across competition writeups.

\subsection*{2. Effective Number of Technologies per Competition}

\begin{figure}[H]
    \centering
    \includegraphics[width=1\linewidth]{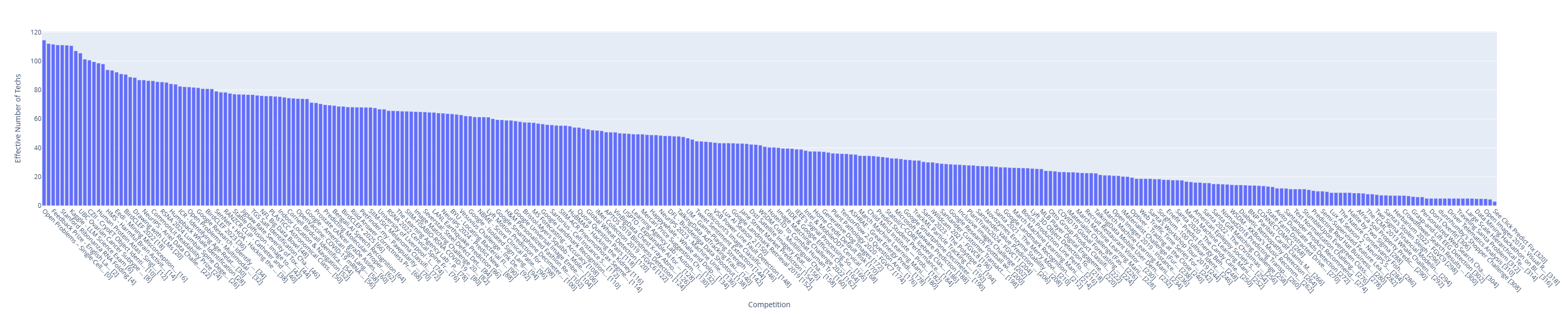}
    \caption{Effective Number of Technologies per Competition \kcite{notebook}{https://www.kaggle.com/code/kevinbnisch/what-technologies-do-winning-kagglers-use}}
    \label{fig:effective-entropy-per-comp}
\end{figure}

Since the normalized entropy can be abstract and unitless, we also compute the \textit{}{effective number of technologies}, defined as:

$$\text{Effective Tech Count} = 2^{H(X)}$$

where $H(X)$ refers to the unnormalized Shannon entropy (in bits). This yields a count-like metric, where--for example--if the technology usage is equivalent to evenly using six different technologies, the effective technology count is 6.  
This interpretation makes diversity comparisons more intuitive, especially in cases where normalized entropy values are tightly clustered.
Figure \ref{fig:effective-entropy-per-comp} shows the results.

We see a different chart now--one that reveals not every competition utilizes an actually broad technology stack, contrary to earlier impressions.  
However, before interpreting these results in detail, we first examine technological diversity aggregated by year, rather than by individual competitions.

\subsection*{3. Technology Diversity Trends by Year}

\begin{figure}[H]
    \centering
    \includegraphics[width=\linewidth]{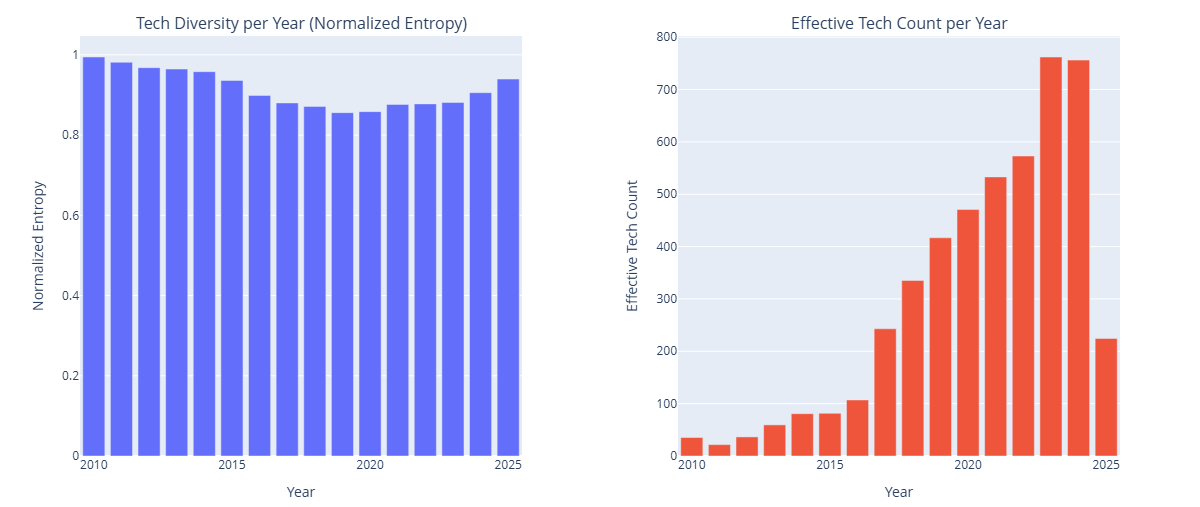}
    \caption{Technology Diversity Trends by Year: Effective vs. Normalized \kcite{notebook}{https://www.kaggle.com/code/kevinbnisch/what-technologies-do-winning-kagglers-use}}
    \label{fig:tech-diversity}
\end{figure}

To analyze how technological diversity has evolved over time, we compare two annual metrics: normalized entropy and the effective number of technologies and plot them through Figure \ref{fig:tech-diversity}.
The normalized entropy per year shows how evenly different technologies were used across competitions in a given year. If the value goes up over time, it means the range of technologies is becoming more diverse. If it stays flat or goes down, it might indicate that fewer technologies are being used more consistently--possibly due to standardization or consolidation.
To again make this easier to interpret, we also show the effective number of technologies per year. This translates the entropy value into a more intuitive number, similar to a count. For instance, an effective count of about \numprint{500} in 2021 means that the mix of technologies used that year was as diverse as if \numprint{500} different technologies were each used equally. This helps us easily spot the years when technological diversity increased or decreased.

\textbf{Why Normalized Entropy and Effective Tech Count Look Different}

The two plots tell different stories because they measure different aspects of technology use.
Normalized entropy shows how evenly technologies are used in a given year. It's a score between 0 and 1. Once usage becomes fairly balanced, the score levels off--even if new technologies are introduced.
Effective tech count, on the other hand, reflects how many technologies were actually used. As more technologies are adopted over time, this number keeps increasing.
In short, the entropy plot tells us that tech usage has stayed balanced, while the effective tech count reveals that the overall variety of technologies has grown significantly.

Overall, we observe that (a) the effective count of technological diversity has been steadily increasing year over year, and (b) the normalized entropy remains consistently high.  
This suggests that Kagglers are not converging on a single dominant technology--although many rely on a common core toolbox--but instead experiment with a variety of approaches and tools.  
Such diversity, often achieved by combining multiple technologies, indicates a broad modeling strategy and aligns well with real-world experimentation and solution development.

\subsubsection{Competition \& Technologies Network}

\begin{figure}
    \centering
    \includegraphics[width=\linewidth]{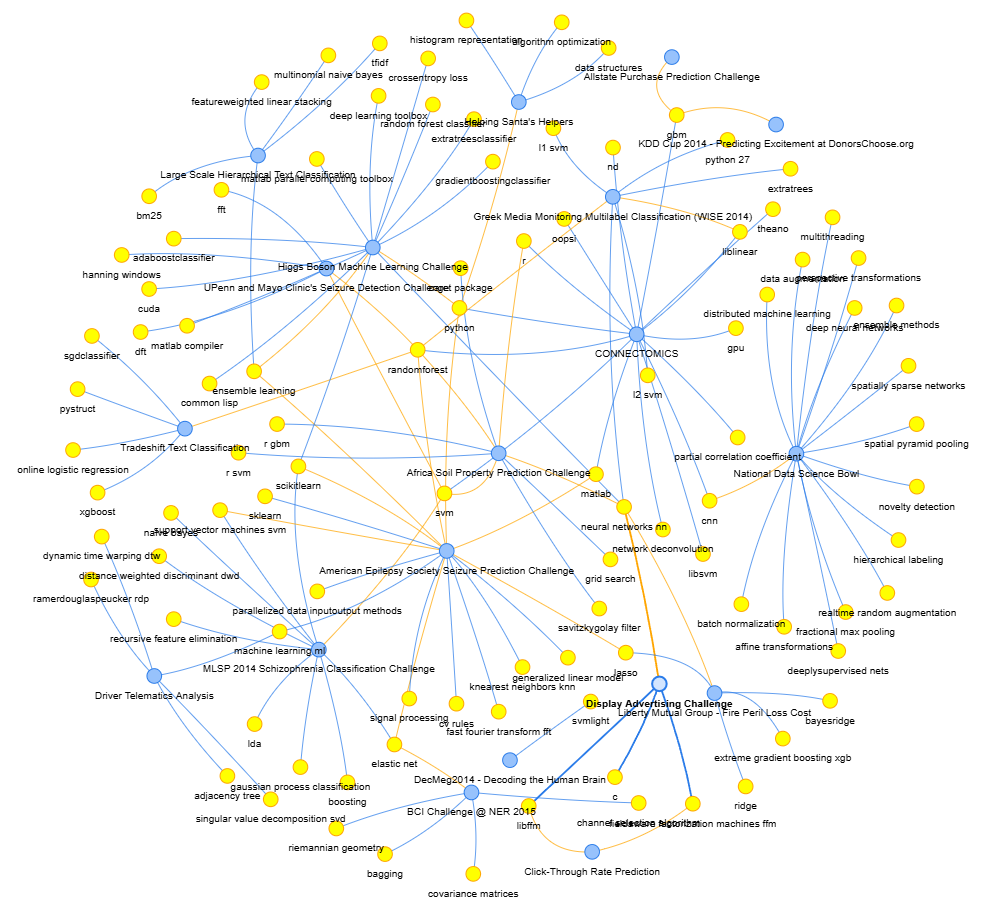}
    \caption{Competition–Technology network for the year 2014.  
    Blue nodes represent competitions, while yellow nodes represent technologies.  
    An edge between a competition and a technology indicates that the technology was mentioned--and therefore used--in that competition's solution write-ups.  
    The color intensity and thickness of each edge reflect the frequency of mentions \kcite{notebook}{https://www.kaggle.com/code/kevinbnisch/what-technologies-do-winning-kagglers-use}.}
    \label{fig:bipartite-network}
\end{figure}

Finally, for each year, we constructed bipartite graphs in which blue-colored nodes represent competitions and yellow-colored nodes represent technologies.  
An edge between a competition and a technology indicates that the technology was either mentioned or used in the corresponding competition.  
Edge thickness and color intensity reflect the frequency of mentions or usage.  
Since these graphs are highly interactive, we include one example graph here as Figure \ref{fig:bipartite-network}--corresponding to the year 2014. For all other yearly graphs and their interactive versions, we refer readers to the accompanying notebook \kcite{notebook}{https://www.kaggle.com/code/kevinbnisch/what-technologies-do-winning-kagglers-use}.

\subsection{Conclusion}

Our analysis of Kaggle kernels reveals clear trends in programming language preferences, technology adoption, and data science practices within the community. 
Python has become the dominant language, now used in approximately 95\% of all kernels, while the use of R has steadily declined since 2016. 

The most frequently imported packages---such as \texttt{pandas}, \texttt{numpy}, and \texttt{matplotlib}---form the backbone of many workflows, underscoring a pragmatic, modeling-driven approach that also values visualization and basic data cleaning.

Over time, we observe increasing entropy and diversity in both package imports and method calls, particularly in competition settings where adaptation to new tools is quick. 
Early dominance by \texttt{xgboost} has shifted toward modern libraries like \texttt{lightgbm}, \texttt{transformers}, and \texttt{autogluon}, with growing interest in AutoML, deep learning, and generative models.

Despite concerns over leaderboard overfitting, our metric-based evaluation across competitions shows that public-private score discrepancies typically range between 5--10\%, suggesting solid generalization in most cases. 

\newpage
\section{Kaggle Competition Topics}\label{sec:meta}
\tldr{
    Kaggle competitions have consistently focused on core machine learning topics like model evaluation, loss functions, and data preprocessing. Over the years, discussions have expanded to include specialized areas such as image generation, autonomous systems, and fraud detection. Despite evolving themes, the community’s focus on practical, performance-driven strategies has remained strong and steadily grown. 
}

In the previous section, we analyzed the \textsc{Meta Kaggle Code} dataset to uncover key insights, trends, and highlight the most significant developments. Within this section, we shift our focus to identifying recurring themes of topics in Kaggle competition forum discussions. As of this writing, approximately 10,000 competitions have been hosted. To minimize our scope we've restricted our analysis to \textit{Featured} and \textit{Research} type competitions with prize pools of at least \$10,000 (or equivalent), yielding a smaller dataset of 418 competitions. \textbf{Topic modeling} is then used to extract the main themes from these discussions.

\subsection{What is Topic Modeling?}

\textbf{Topic modeling} is a \textbf{Natural Language Processing} (NLP) technique used to identify underlying themes within a large collection of text by grouping frequently co-occurring words into meaningful topics \cite{aly-2023-topic}. These clusters of terms naturally form patterns that reveal the main topics or themes of topics being discussed. While traditional approaches like \textbf{Latent Dirichlet Allocation (LDA)} \cite{blei-2003-latent} and \textbf{Latent Semantic Analysis (LSA)} \cite{deerwester1990latent} have been widely used. Recent advancements in technology and processing power have paved way for newer, more sophisticated techniques. In this section, we will be utilizing \textbf{BERTopic} \cite{grootendorst-2022-bertopicneuraltopicmodeling}, which, under-the-hood, uses pre-trained transformer models such as BERT to generate contextual embeddings for improved topic extraction.

Compared to classical methods, \textbf{BERTopic} offers greater accuracy and a deeper understanding of context, particularly when working with short or complex texts. Its flexibility and interpretability make it especially valuable for large-scale text analysis across various domains, including business intelligence, social media monitoring, and customer feedback analysis. 

\subsection{Topic Modeling Kaggle Competitions}

\begin{figure}
    \centering
    \includegraphics[width=\linewidth]{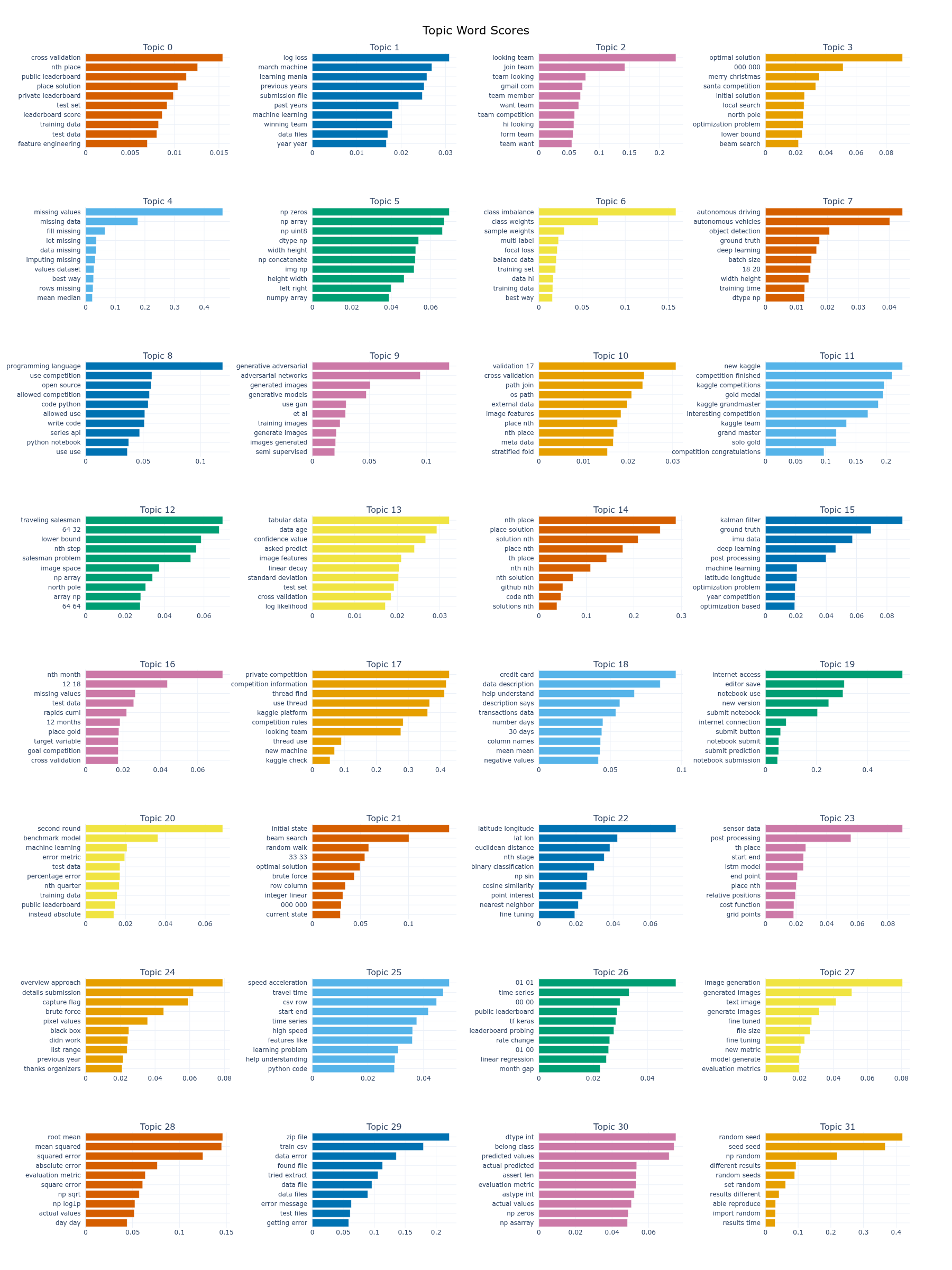}
    \caption{Competition Topics and Associated Most Important Terms\kcite{notebook}{https://www.kaggle.com/code/bwandowando/generate-bertopic-competitions-only}}
    \label{fig:comp-topics-and-terms}
\end{figure}

Based on the \textbf{BERTopic} model applied to Kaggle \textit{Featured} and \textit{Research} competitions with prize pools of at least \$10,000, we observe several key topics that are being discussed by participants in multiple competitions as shown in Figure \ref{fig:comp-topics-and-terms}. Across all competitions, the most dominant topic, \textbf{Topic 0}, centers around model evaluation and feature engineering. Terms like \enquote{cross validation}, \enquote{public leaderboard}, \enquote{private leaderboard} and \enquote{feature engineering} highlight how participants prioritize understanding of how their models are scored, generalize to unseen data, and rank on leaderboards. This reflects the competitive nature of Kaggle participants, where effective validation strategies and well-engineered features are critical for achieving top rankings.

Following this, \textbf{Topic 1} focuses on loss functions and model performance, with key terms such as \enquote{log loss}, \enquote{mean average precision}, and \enquote{winning rate} indicating discussions around optimizing evaluation metrics specific to competition goals. \textbf{Topic 2} reflects team collaboration, with phrases like \enquote{team member}, \enquote{looking team} and \enquote{team work} showing the importance of group effort and shared insights. \textbf{Topic 3} is centered on optimization problems, involving terms like \enquote{optimal solution}, \enquote{lower bound} and \enquote{optimization problem} suggesting that participants often discuss theoretical and practical approaches to improving model performance. \textbf{Topic 4} is largely about missing data and imputation, with high scores for words like \enquote{missing values} \enquote{impute} and \enquote{NA values} indicating a strong focus on data preprocessing as a foundational step in many competitions.

Beyond the top five, the remaining topics cover a wide range of themes relevant to machine learning competitions. \textbf{Topic 6} addresses \enquote{class imbalance}, a common challenge in classification tasks. \textbf{Topics 7 and 27} discuss \enquote{autonomous systems} and \enquote{image generation}, respectively, pointing to the presence of specialized domains such as computer vision and autonomous driving. \textbf{Topics 10 and 11} focus on \enquote{competition logistics} and \enquote{platform features}, reflecting discussions about how competitions are run and participant engagement.

In general, these topics show that while core modeling and data strategies occur the most, participants also engage deeply with on competition-specific discussions.

\begin{figure}
    \centering
    \includegraphics[width=\linewidth]{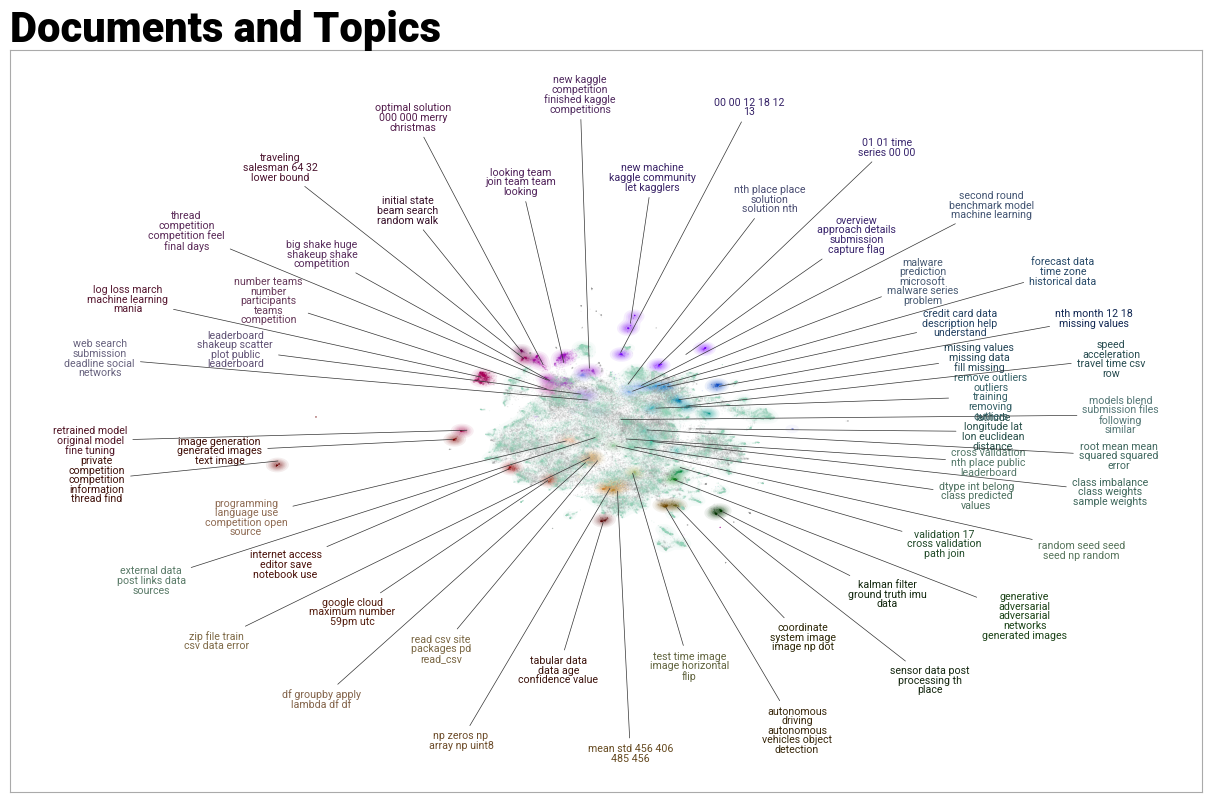}
    \caption{Competition Topics Data Map\kcite{notebook}{https://www.kaggle.com/code/bwandowando/generate-bertopic-competitions-only}}
    \label{fig:comp-topic-data-map}
\end{figure}

The graph displayed in Figure \ref{fig:comp-topic-data-map} contains the same set of topics as outlined before, but presented in a 2D visualization of topic clusters. We generated this graph the built in function \texttt{visualize\_document\_datamap()} of a BERTopic instance. Each point represents a document or post from the dataset, and points that are closer together share higher semantic similarity, meaning they discuss similar ideas or themes.

Topics are shown as color-coded clusters, and their spatial distribution reflects how near or distant the topics are from one another. For example, closely grouped labels such as \enquote{log loss}, \enquote{public leaderboard} and \enquote{cross validation} occupy the central area, indicating these themes are core to most discussions and overlap significantly across many competitions. In contrast, more specialized topics like \enquote{image generation}, \enquote{autonomous driving} and \enquote{credit card fraud} appear on the outer edges, representing niche domains or task-specific discussions.

The visualization helps us understand not only what topics are discussed, but how interconnected they are. General machine learning principles such as model evaluation, data preprocessing, and leaderboard strategies form a dense central cluster, while more technical or domain-specific topics spread outward, illustrating the breadth of conversations in high-stakes Kaggle competitions.

\begin{figure}
    \centering
    \includegraphics[width=\linewidth]{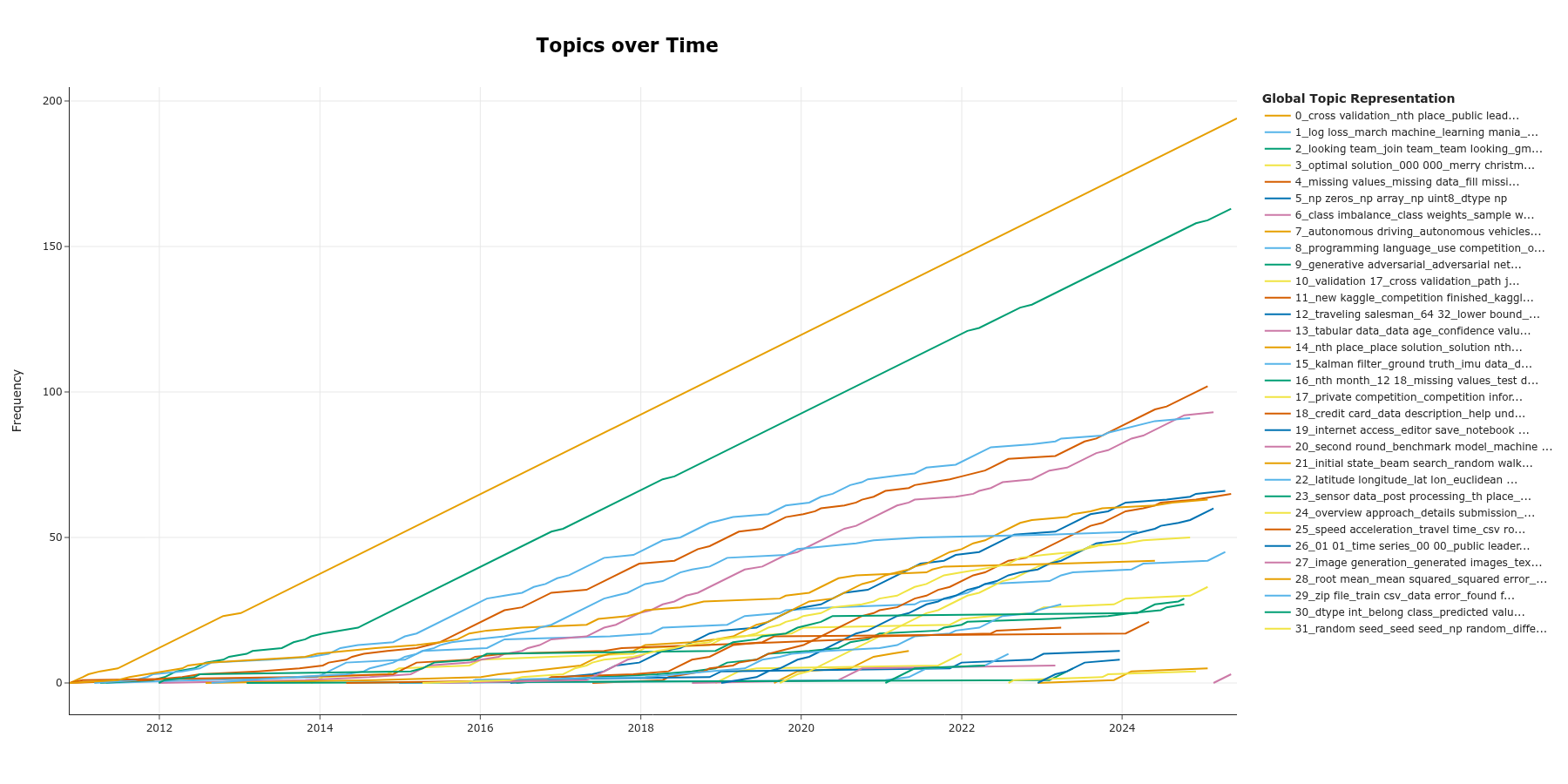}
    \caption{Competition Topics Over Time\kcite{notebook}{https://www.kaggle.com/code/bwandowando/generate-bertopic-competitions-only}}
    \label{fig:comp-topics-over-time}
\end{figure}

Figure \ref{fig:comp-topics-over-time} shows how often different topics were discussed in Kaggle competitions from 2011 to 2025. It highlights that while new topics have appeared over the years, many key themes have stayed relevant and consistent. The steady growth of most topics over the years suggests that interest in both general machine learning concepts and specific problem areas has continued to grow over time. Also, the number of members that are joining Kaggle and participating in competitions is growing as well.

From early years through to the most recent competitions, the diversity of discussions has broadened, indicating an expanding scope of challenges and methodologies in the Kaggle ecosystem. Some topics gained momentum quickly and sustained high relevance, while others appeared later but grew rapidly, reflecting evolving interests in areas such as deep learning, specialized domains, and practical implementation issues. Overall, the chart highlights the evolving, yet stable nature of knowledge exchanges in discussions within Kaggle competition forums.

\subsection{Topic Modeling Specific Competitions}

In the previous section, we identified recurring themes across multiple Kaggle competitions. To explore whether these patterns also emerge at the individual competition level, we selected one of Kaggle's competitions with the highest number of participants. Generally, competitions with a larger number of participants tend to generate more forum activity, resulting in a greater volume and variety of discussion threads.

{\href{https://www.kaggle.com/competitions/santander-customer-transaction-prediction}{\textbf{The Santander Customer Transaction Prediction}}} competition challenged participants to build models that predict which customers will make a specific future transaction, focusing on a binary classification task. With 25,500 entrants, over 104,000 submissions, and a \$65,000 prize pool, it became one of Kaggle's most participated competitions. Hosted by {\href{https://www.santanderbank.com/}{Santander Bank}}, the competition offered anonymized, real-world-like data to encourage innovative solutions from the global data science community.

\begin{figure}
    \centering
    \includegraphics[width=\linewidth]{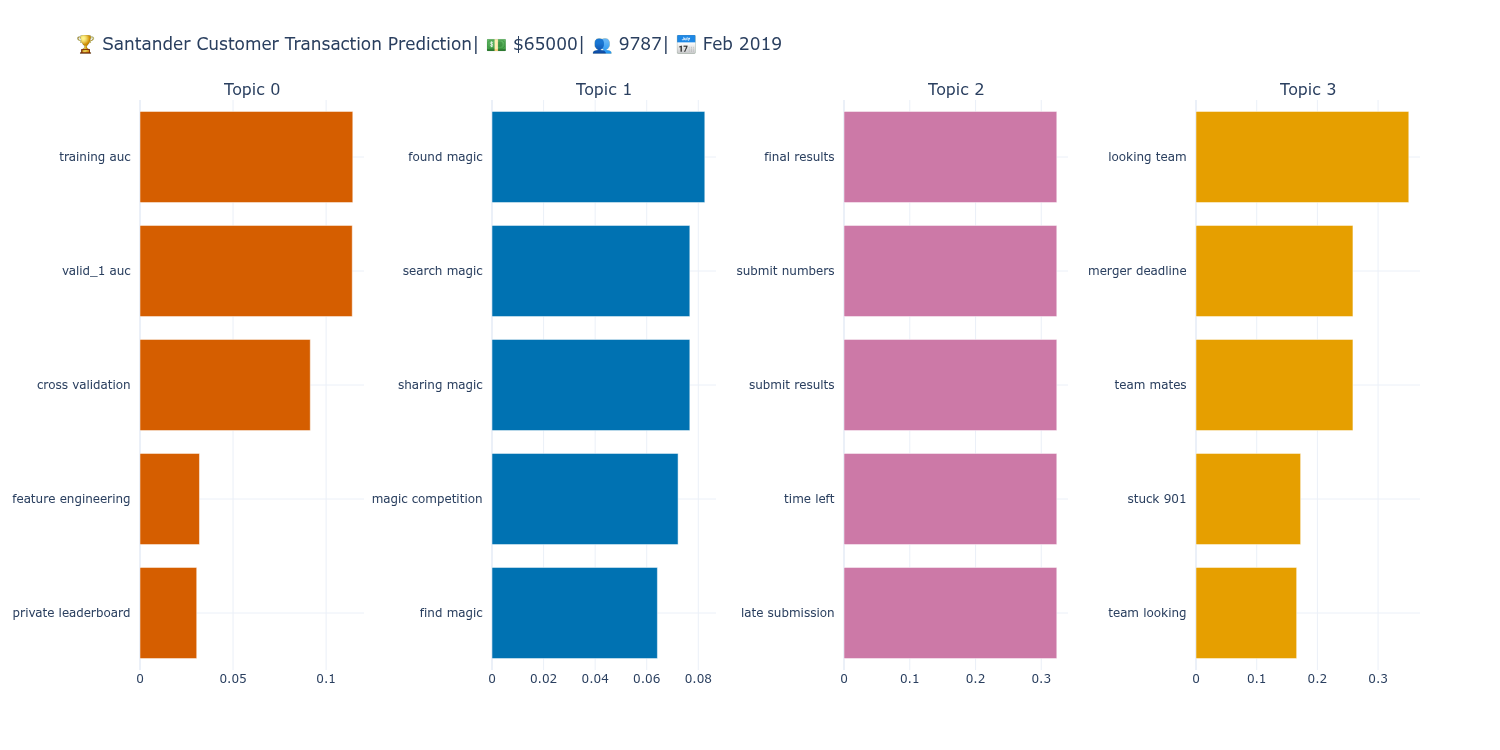}
    \caption{Santander Topics and Associated Most Important Terms\kcite{notebook}{https://www.kaggle.com/code/bwandowando/bertopics-of-top-20-competitions-most-with-users}}
    \label{fig:santander-barchart}
\end{figure}

We extracted all the topics that were created during the competition was active and up to a month after the competition has closed. Figure \ref{fig:santander-barchart} is a topic barchart of the competition showing most important terms (words) associated with each topic, ranked by their c-TF-IDF scores. 

\begin{figure}
    \centering
    \includegraphics[width=\linewidth]{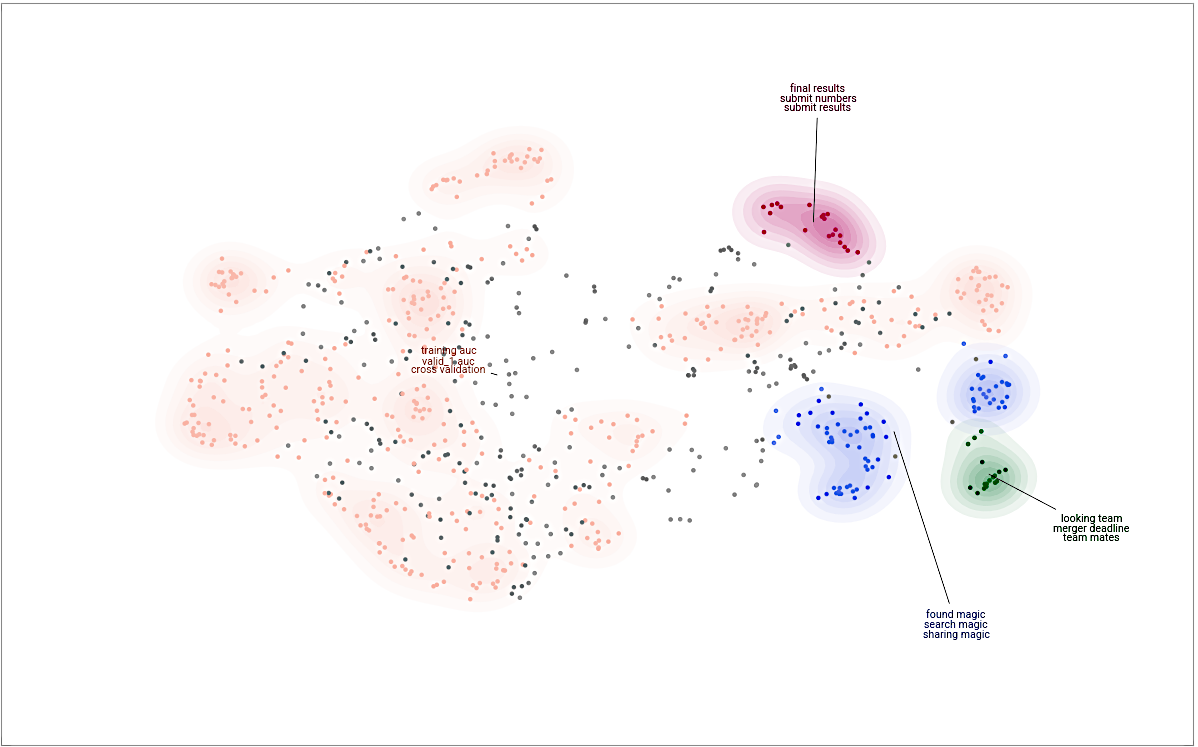}
    \caption{Santander Topics Data Map\kcite{notebook}{https://www.kaggle.com/code/bwandowando/bertopics-of-top-20-competitions-most-with-users}}
    \label{fig:santander-data-map}
\end{figure}

Figure \ref{fig:santander-data-map} visualizes the clustering of the topics' embeddings and we can definitely see that most of the discussions revolve around training and test data, cross-validation strategies and the metric being used in the competition. And based on the discussion, we get to know that the competition's chosen metric is AUC metric.

\begin{figure}
    \centering
    \includegraphics[width=\linewidth]{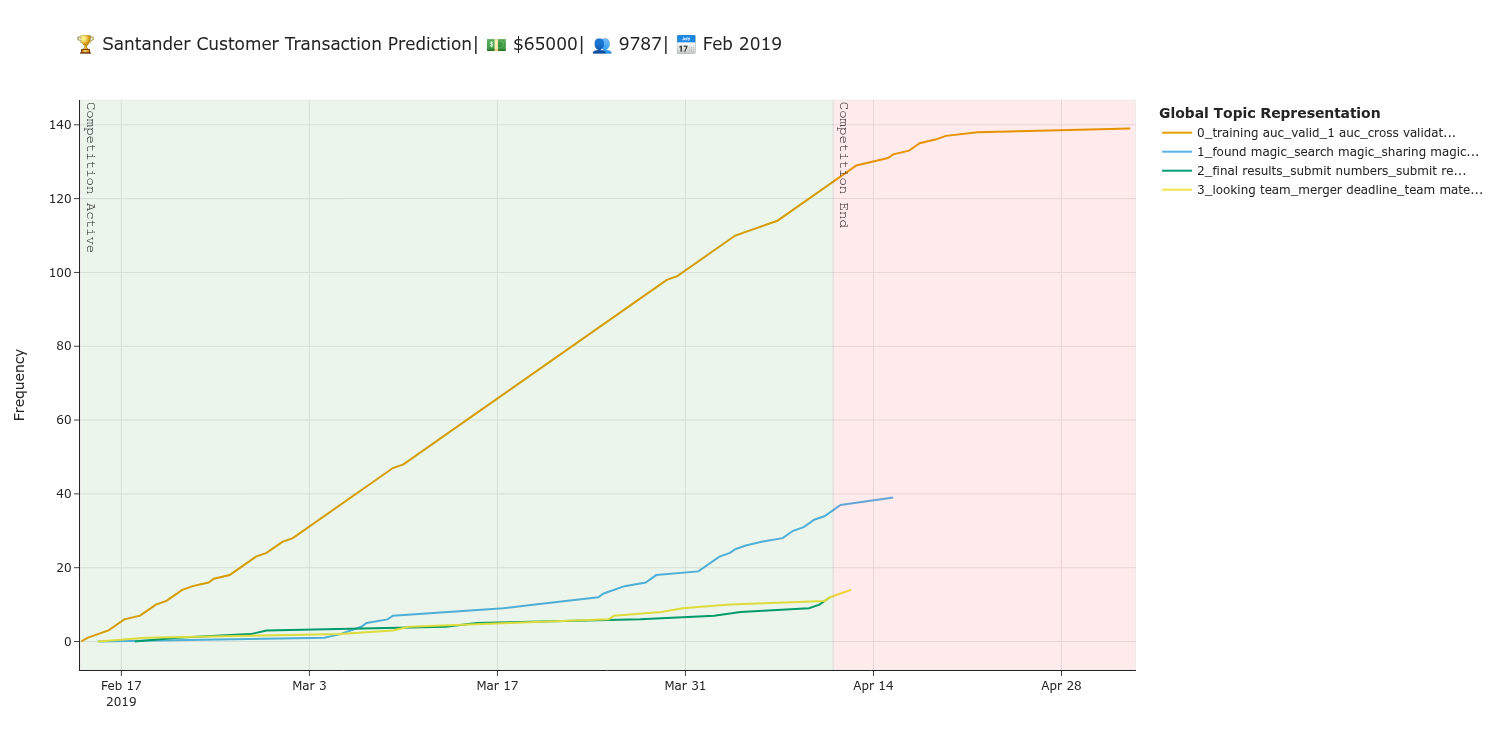}
    \caption{Santander Topics over Time\kcite{notebook}{https://www.kaggle.com/code/bwandowando/bertopics-of-top-20-competitions-most-with-users}}
    \label{fig:santander-topcs-over-time}
\end{figure}

Figure \ref{fig:santander-topcs-over-time} visualizes the evolution of discussion topics in the Kaggle forum for the "Santander Customer Transaction Prediction" competition, which began in February 2019. The graph tracks cumulative frequencies of four main topic clusters over time, with the competition period shaded in green and the post-competition period in red. During the competition, the most dominant topic was related to model training and cross-validation strategies, reflecting participants’ focus on improving performance. 

\subsection{Conclusion}

The results confirm that the same key topics identified across all competitions, model evaluation, leaderboard strategies, and feature engineering, are also consistently discussed within individual competitions. These topics discuss the fundamentals which participants need to adhere to become effective when competing in Kaggle competitions.

The analysis of Kaggle competitions reveals consistent engagement with fundmanental machine learning topics and techniques such as model evaluation, leaderboard strategies, and data preprocessing. These themes have remained central across the years, reflecting their importance in achieving competitive results. At the same time, the growth of domain-specific topics including image generation, autonomous systems, and fraud detection, demonstrates that both the platform and community evolves and adapts to real-world challenges and industry trends. The visual distribution of topics shows clear clustering around core technical discussions, with more specialized topics forming distinct but related areas. Our analysis further highlights the sustained and increasing interest in both general and niche topics.

In general, the findings underscore the maturity and depth of the Kaggle community, as well as its ongoing collective effort to pursue high-quality, solution-driven collaboration in data science.
\newpage

\bibliographystyle{plain}
\bibliography{references} 

\section*{Appendix}

\end{document}